\pdfoutput=1

\documentclass[11pt]{article}

\usepackage[]{ACL2023}

\usepackage{times}
\usepackage{latexsym}
\usepackage{graphicx}
\usepackage{colortbl}
\usepackage{float}
\usepackage{subcaption}
\usepackage{booktabs}

\definecolor{InterviewSetting}{HTML}{e5b0b0}
\definecolor{HumanExperience}{HTML}{8dd986}
\definecolor{SubjectiveInterpretation}{HTML}{f7b182}
\definecolor{ObjectiveInterpretation}{HTML}{bd70fe}
\definecolor{Linguistic}{HTML}{7d94fe}

\usepackage{multirow}
\usepackage[normalem]{ulem}
\useunder{\uline}{\ul}{}
\usepackage{lscape}

\usepackage{graphicx}     
\usepackage{array}        
\usepackage{lipsum}       
\usepackage{multicol}     
\usepackage{longtable}    

\usepackage[T1]{fontenc}

\usepackage[utf8]{inputenc}

\usepackage{microtype}

\usepackage{inconsolata}
\usepackage{amssymb}
 \usepackage{amsmath}
 \usepackage[capitalize,noabbrev]{cleveref}

\definecolor{bblue}{rgb}{0.0, 0.6, 0.8}

\title{Dementia Through Different Eyes: Explainable Modeling of Human and LLM Perceptions for Early Awareness}

\author{
    Lotem Peled-Cohen\textsuperscript{*  T} \quad
    Maya Zadok\textsuperscript{*  T} \quad
    Nitay Calderon\textsuperscript{ T} \quad
    Hila Gonen\textsuperscript{ U} \quad
    Roi Reichart\textsuperscript{ T} \\ 
    \textsuperscript{T }Faculty of Data and Decision Sciences, Technion \\
    \textsuperscript{U }Paul G. Allen School of Computer Science \& Engineering, University of Washington \\
    \texttt{\{splotem,maya.zadok,nitay\}} @ \texttt{campus.technion.ac.il}\\
    \texttt{hilagnn@gmail.com} \quad
    \texttt{roiri@technion.ac.il}
}

\begin{document}
\maketitle
\begin{abstract}

\def\thefootnote{*}\footnotetext{Equal contribution.}

Cognitive decline often surfaces in language years before diagnosis. It is frequently non-experts, such as those closest to the patient, who first sense a change and raise concern. As LLMs become integrated into daily communication and used over prolonged periods, it may even be an LLM that notices something is off. But what exactly do they notice--and \textit{should be noticing}--when making that judgment? This paper investigates how dementia is perceived through language by non-experts. We presented transcribed picture descriptions to non-expert humans and LLMs, asking them to intuitively judge whether each text was produced by someone healthy or with dementia. We introduce an explainable method that uses LLMs to extract high-level, expert-guided features representing these picture descriptions, and use logistic regression to model human and LLM perceptions and compare with clinical diagnoses. Our analysis reveals that human perception of dementia is inconsistent and relies on a narrow, and sometimes misleading, set of cues. LLMs, by contrast, draw on a richer, more nuanced feature set that aligns more closely with clinical patterns. Still, both groups show a tendency toward false negatives, frequently overlooking dementia cases. Through our interpretable framework and the insights it provides, we hope to help non-experts better recognize the linguistic signs that matter.

\end{abstract}

\section{Introduction}

Dementia is a progressive neurodegenerative condition caused by various underlying pathologies, most commonly Alzheimer’s disease (AD). Tens of millions are currently living with the disease worldwide, with this figure expected to double with each passing generation \cite{alzintStats2025}. Early diagnosis is critical for maximizing the effectiveness of both symptomatic and disease-modifying interventions, especially during mild cognitive impairment (MCI), a transitional stage between normal aging and dementia, where cognitive deficits are present but do not yet impair daily functioning \citep{prince2011world, rasmussen2019alzheimer, simsDonanemabEarlySymptomatic2023}.

Subtle language dysfunctions have long been recognized as early signs of cognitive decline, making linguistic signals valuable for early detection \citep{verma2012semantic, szatloczki2015speaking, orimaye2017predicting, martinez2021ten, cho2022lexical}. This has spurred hundreds of studies applying Natural Language Processing (NLP) methods to transcribed cognitive assessments \citep{peled2024systematic}, primarily to extract linguistic markers (disfluencies, rephrasing, part-of-speech ratios, etc; \citealp[]{soni2021using, adhikari2021comparative, farzana-etal-2022-say, williams-etal-2021-relationships}) and to detect cognitive decline.

In reality, however, the initial ``detection'' of dementia symptoms rarely begins with a clinician or structured cognitive assessments. Instead, it is often the individuals themselves or their close environment, who first notice signs of cognitive decline and initiates a medical evaluation \citep{van2018subjective, scharre2019preclinical, jessen2020characterisation}. This highlights the importance of understanding how dementia is perceived by non-experts, i.e., those who are not yet patients, caregivers, or clinicians. Identifying which linguistic behaviors are perceived as related to dementia can reveal where public intuition aligns with clinical insight, and where it falls short and can benefit from further education. 

Additionally, nowadays, it is not just humans who can track linguistic changes and raise concerns. Adults over 55, an age group at increased risk for cognitive decline \cite{bai2022worldwide}, now regularly use large language models (LLMs)  \cite{retirementlivingFutureSenior}. Given these models' extensive world knowledge and continuous linguistic signal from the user, one can imagine
LLMs flagging subtle shifts in language to indicate early signs of cognitive impairment. It is therefore important to understand which cues drive LLMs' perceptions.

Our paper explores the concept of \textit{dementia perception by non-experts}: which cues lead humans and LLMs to perceive someone as cognitively impaired based on their language. To study this, we collected perceptions from 27 humans and 3 LLMs--LLaMA 3 \cite{grattafiori2024llama}, GPT-4o \cite{achiam2023gpt}, and Gemini-1.5-Pro \cite{team2024gemini}--who were presented with 514 transcriptions of a spontaneous speech task (Cookie Theft picture descriptions from the Pitt corpus; \citealp{becker1994natural}; further detailed in \cref{sub:background_cookie}). Humans and LLMs were asked to give their best intuitive judgment as to whether each text was produced by someone healthy or cognitively impaired. Throughout this work, we use ``humans'' to refer to non-expert humans, and ``clinicians'' or ``clinical diagnosis'' to refer to expert medical judgment.

To analyze perceptions, we propose a 4-step explainable method, inspired by studies such as \citet{badian2023social}, \citet{balek2024llm}, and \citet{lissak2024bored}:
(1) design intuitive, human-centered features in consultation with domain experts;
(2) extract these features using an LLM as an annotator, with quality control;
(3) train inherently simple and interpretable logistic regression models to predict perceptions and clinical diagnosis; and
(4) analyze coefficients to identify the linguistic cues influencing how dementia is perceived. \cref{sec:perception_annotations,sec:feature_ex_method,sec:logreg_modeling,sec:results} outline the full method and results. 

Our pipeline is rooted in high-level features that capture nuanced aspects of picture descriptions. These 38 binary features (\cref{tab:full_features}, \cref{appendix:full_feature_list}), were developed in accordance with established literature and in collaboration with domain experts. The features span five categories aligned with cognitive processes involved in describing a picture: visual processing (e.g., ``I see a boy and a girl'');  reasoning (e.g., ``he it about to fall''); verbal expression (e.g., disfluencies); emotional reaction (e.g., ``poor kids''); and personal interactions (e.g., ``Is that what you meant?''). This categorization allows for an analysis beyond individual features, and conclusions that may generalize beyond picture descriptions to other texts produced by patients. 

Traditionally, extracting such high-level features would require extensive feature engineering and a dedicated algorithmic logic for each feature, posing a significant scalability challenge. Manual annotation is also impractical, as our dataset includes 38 features for hundreds of transcriptions, amounting to over 19,000 annotations. To address this, we use LLMs as annotators and validate their output with statistical tests to ensure quality comparable to human annotations \cite{calderon2025alternative}.



Our analysis reveals that human judgments are highly inconsistent and show a tendency toward false negatives, i.e., labeling clinically diagnoses cases as healthy. Humans appear to rely on a narrow set of simple, objective features, and sometimes interpret cues in ways that contradict clinical patterns. Notably, when asked to describe which linguistic cues shaped their perception, they often reported features that align with our predefined set, reinforcing its validity. However, these self-reports do not match the actual decision patterns, suggesting that people do not rely on what they think they do. LLMs, while also prone to false negatives, appear to rely on a much richer feature set, including sentiment-related cues. This highlights their surprisingly nuanced use of language in this task.



To summarize, our contributions are as follows: (1) we propose an interpretable four-step method building on LLMs-as-annotators with statistical significance quality test for analyzing the linguistic features driving dementia perception; (2) we analyze the perceptions of non-experts (27 human annotators and three LLMs) and identify the linguistic behaviors they associate with dementia; (3) we examine the extent to which these behaviors overlap with features associated with clinical diagnosis and analyse human and LLM misperceptions. 

We hope this study lays a foundation for future research on dementia perception from the perspective of all stakeholders. By shedding light on the linguistic cues that non-experts and LLMs rely on when assessing cognitive decline, we believe our findings can contribute to broader public awareness and support earlier detection. Finally, we aspire to encourage the interpretable and statistically grounded use of LLMs in sensitive domains, fostering interdisciplinary trust and real-world impact.

\section{Related Work}
\label{related_work}

\textbf{NLP-Based Dementia Detection}\quad NLP is increasingly used in dementia research, typically applied to transcribed clinical assessments to detect cognitive decline or identify its linguistic markers \cite{peled2024systematic}. Commonly used datasets include the CCC corpus \cite{pope2011finding}, the ADReSS and ADReSSo challenge sets \cite{luz2021detecting,luz2021alzheimer}, and the popular Pitt corpus (\citealp{becker1994natural}; detailed in \cref{sub:background_cookie}). All four datasets provide transcribed speech from both healthy and cognitively impaired individuals.

\begin{figure*}[t]
    \centering
    \includegraphics[width=0.9\linewidth]{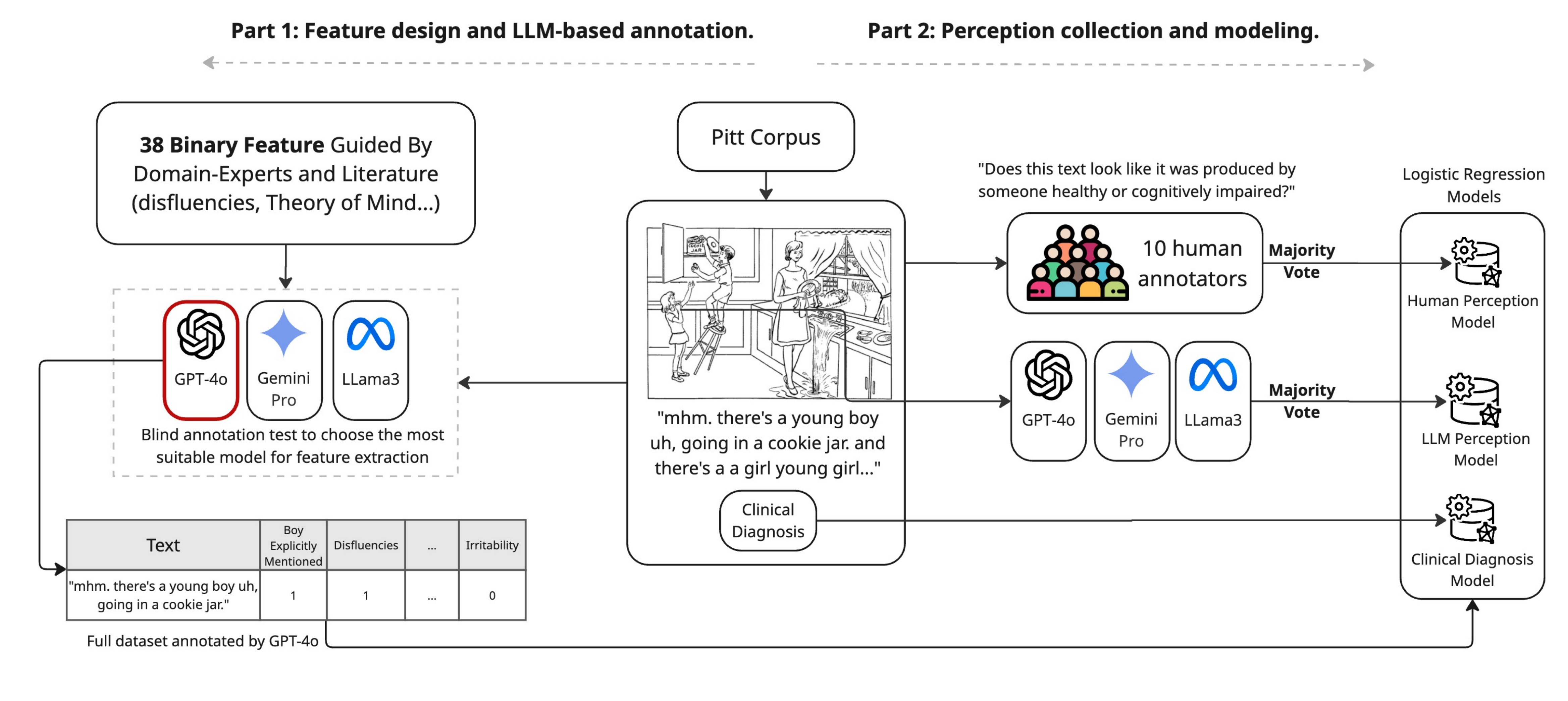}
    \caption{Illustration of the end-to-end methodological process.}
    \label{fig:methodology_flowchart}
    \vspace{-0.5em}
\end{figure*}

These datasets have been used to extract linguistic markers such as syntactic complexity \cite{roark2007syntactic}, idea density \cite{sirts2017idea}, topic structure \cite{pompili2020pragmatic}, and meta-semantic terms, i.e., words expressing emotion or opinion \cite{choi2019meta}. Disfluencies, pauses, and context shifts are other markers known to significantly influence model predictions \cite{kemper1989language, adhikari2021comparative, farzana2022you}. Markers are then used to train dementia detection algorithms, using traditional classifiers such as SVMs and Random Forests \citep{jarrold2014aided, fraser2016linguistic, zhou2016speech} or transformers \citep{pappagari2020using, edwards2020multiscale,  balagopalan2021comparing}. Recently, LLMs have also been leveraged for feature extraction or embedding-based representations supporting dementia detection \cite{li2023two, liu2023leveraging, bang2024alzheimer, botelho2024macro, bt2024performance, koga2024evaluating, latif2024evaluation, runde2024optimization}. Despite these advances, automated detection has yet to be adopted in clinical settings, and early diagnosis still relies on self-reports from patients or their environment.

\textbf{Explainable Dementia Detection}\quad In high-risk fields such as healthcare, explainability is crucial for establishing trust and encouraging the adoption of artificial intelligence systems \citep{adadi2020explainable}. A wide range of interpretability paradigms is available to NLP practitioners \citep{stiglic2020interpretability, calderon2024behalf, viswan2024explainable}, 
and some dementia-related studies do leverage them. \citet{karlekar2018detecting} analyzed model predictions using activation clustering and derivative saliency, while \citet{ilias2022explainable} examined text statistics using LIME \cite{ribeiro2016should} to identify which words or phrases most influenced individual predictions. Others, such as \citet{vimbi2024interpreting}, use feature attribution methods that assign importance to individual features or tokens. While related in spirit, our approach goes further by attributing predictions to higher-level, cognitively grounded concepts, offering more interpretable and generalizable insights than assigning importance to raw inputs like tokens.


\textbf{Societal Perception of Dementia}\quad Studies on how dementia is perceived by the general public primarily focus on the concept of stigma. Long recognized as a defining aspect of the dementia experience \citep{graham2003reducing, milne2010d, benbow2012dementia}, stigma has been shown to negatively impact emotional well-being and contribute to delays in diagnosis and help-seeking \citep{swaffer2014dementia, gove2016stigma,  nguyen2020understanding}.

Few studies have used NLP to investigate this topic, and those that have focus almost exclusively on dementia portrayal in social media conversations \citep{oscar2017machine, pilozzi2020overcoming, tahami2022stakeholder}, revealing that patients are mocked or ridiculed. While social media is accessible and captures real-world language, it has key limitations: (1) scarce presence of dementia patients \cite{panzavolta2025patient}, making their language hard to study; (2) content is often pre-planned or edited; and (3) unlike the clinical data we use, social media data is not centered on cognitive abilities impacted by dementia, or designed to elicit narrative, reasoning, or spontaneous dialogue.

\textbf{\textit{Our Novelty:}} To the best of our knowledge, this is the first work to examine how non-experts and LLMs perceive dementia through clinical cognitive assessments. We uniquely approach the task by introducing an interpretable method rooted in human-oriented, expert-guided high-level features, extracted using LLMs as annotators, and used to model perception. Unlike prior work, we do not focus on the prediction of clinical diagnosis but rather on understanding the non-experts: their reasoning, where their perceptions diverge from clinical judgment, and how intuition may be improved. 


\section{Cookie Theft Picture Descriptions}
\label{sub:cookie_theft}

\subsection{Background}
\label{sub:background_cookie}


In this section, we describe the data used to obtain and analyze perceptions. We rely on the Pitt corpus \cite{becker1994natural}, a widely used dataset from the DementiaBank cohort \cite{lanzi2023dementiabank}, which includes longitudinal data from 98 healthy individuals and 196 dementia patients across varying stages and types, primarily AD. The corpus provides participant demographics, standardized cognitive scores and transcribed recordings of various linguistic tasks, such as the Cookie Theft picture description, which we focus on in our study.

Originally designed by \citet{goodglassBDAEBostonDiagnostic1983}, the Cookie Theft picture description task is commonly used to elicit spontaneous speech in cognitive assessments  \cite{gorenExpressiveLanguageCharacteristics1992, fraserLinguisticFeaturesIdentify2016, berubeAnalysisRightHemisphere2022, butalaParkinsonicsRandomizedBlinded2022}. In this task, participants are shown an image of a domestic scene (\cref{fig:cookie}) and asked to describe what they see, with their responses typically recorded, transcribed, analyzed, and scored. However, this task provides clinicians with more than just standardized scores, as they commonly use it to form \textit{a general impression} of an individual’s information processing, linguistic performance, motor speech function, and communicative ability. The fact that clinicians routinely rely on this task to make such holistic judgments highlights its practical value for our study of perception and intuition. 



\subsection{Preprocessing}
\label{sub:preprocessing_cookie}
We extract 514 Cookie Theft Picture descriptions and their corresponding clinical diagnosis (`Healthy Control', `MCI', `Possible AD', `Probable AD', `AD', and `Other'). We then binarize these into two classes: `Healthy' and `Dementia', with the latter including all samples not labeled as `Healthy Control'. This binary framing simplifies the task for non-expert annotators, since differentiating between the types and stages of cognitive decline is challenging even for trained clinicians.


As mentioned in \cref{related_work}, this corpus has been extensively used in NLP-focused dementia research. Building on these studies, we apply standard preprocessing to the Cookie Theft transcripts, separating the participant’s speech from the interviewer’s and removing extraneous characters and interview-specific annotations. For the remainder of the study, the input provided to both humans and LLMs consists solely of raw transcriptions, with no additional demographic information or clinical labels. This helps mitigate potential biases, such as assumptions based on the speaker’s age, while also complying with the dataset's safety regulations.


\section{Perception Annotations}
\label{sec:perception_annotations}

\subsection{Human Perception} We recruited 27 non-expert annotators to read the preprocessed picture descriptions and intuitively judge each as ``Healthy'' or ``Dementia''. Annotators received no prior instructions regarding which cues to consider when making their decisions (see \cref{app:annotation_guidelines} for annotation guidelines). Generally, we focused on young non-expert adults, none have prior experience as dementia caregivers. Full demographic details about our annotators are presented in \cref{tab:human_annotators}. We deliberately chose not to include other populations, such as caregivers or clinicians, as (a) expert and non-expert groups are likely to produce significantly different perception signals, prompting dedicated studies; and (b) young adults are a particularly relevant group for studying dementia perception, as they are increasingly likely to observe early signs of MCI in their close circles. Understanding what they perceive, and eventually helping them recognize those signs, is crucial.



Each description was labeled by 10 annotators, and the majority vote was used as the final perception label. No ties occurred in any of the samples. Annotations show a relatively low inter-annotator agreement (Fleiss’ $\kappa = 0.28$), which is unsurprising given the inherent subjectivity of the task \cite{rottger2022two}. After completing the task, annotators were asked to describe any cues they noticed that may have influenced their decision.

\begin{figure}[t]
    \centering
    \includegraphics[width=1\linewidth]{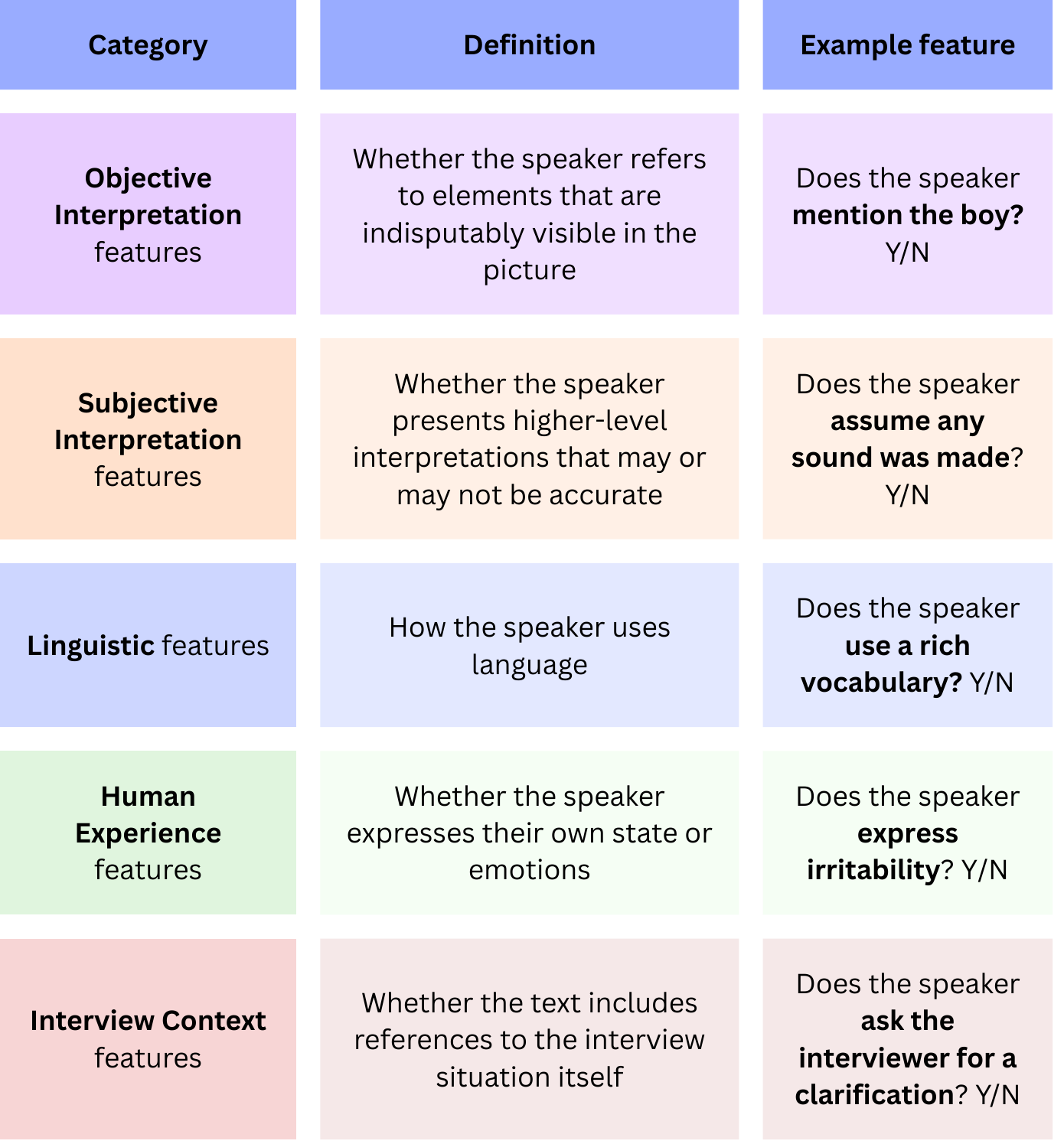}
    \caption{Feature categories, definitions, and examples.}
    \label{fig:feature_categories}
    \vspace{-0.5em}
\end{figure}

\subsection{LLM Perception} GPT-4o, LLaMA 3, and Gemini-1.5-Pro were provided with the same transcripts and also asked to provide their best judgment (see Appendix~\ref{app:llm_perception_prompt} for the full prompt). Each model labeled the entire dataset, and we used their majority vote in our analysis. LLMs show stronger inter-annotator agreement than humans (Fleiss’ $\kappa = 0.465$), perhaps due to their shared training data and structure.

\section{Feature Extraction Methodology}
\label{sec:feature_ex_method}


\subsection{Feature Design} 


Our core intuition is that when a text is perceived as produced by someone with dementia, judgment is not based on computational scores such as noun-to-verb ratio  \cite{williamsLexicalsemanticPropertiesVerbs2023}. Instead, it often stems from a gut feeling--whether the text feels informative or empty, comprised of rich or repetitive vocabulary, etc. We therefore aim to define features in a more intuitive manner. For example, to represent the noun-to-verb ratio, one can ask: ``Did the speaker focus on actions over objects?''. This framing makes the feature easier to interpret and potentially adopt as a guideline.

In collaboration with a neurologist and a neuropsychologist, we defined \textbf{38 binary features} that capture informative aspects of the picture descriptions in an intuitive manner, and are anchored in dementia research (see \cref{tab:full_features},  \cref{appendix:full_feature_list} for all features, sources and examples). The features span five categories (Figure~\ref{fig:feature_categories}) aligned with cognitive processes involved in picture description: what is directly observed  (\textbf{\textcolor{ObjectiveInterpretation}{Objective Interpretation}}); what is inferred or assumed beyond what is seen (\textbf{\textcolor{SubjectiveInterpretation}{Subjective Interpretation}}); how these observations are linguistically expressed (\textbf{\textcolor{Linguistic}{Linguistic}}); and the emotional or experiential states expressed throughout (\textbf{\textcolor{HumanExperience}{Human Experience}} and \textbf{\textcolor{InterviewSetting}{Interview Context}}). Grouping features into broader categories enables higher-level analysis and more generalizable insights; \citet{cummings2019describing}, for example, found that related feature groups may be more reliable diagnostic markers than individual cues.



\subsection{Feature Annotation}

Building on studies such as \citet{he2023annollm, tan2024large} and \citet{badian2023social}, we use LLMs as annotators.  We experiment with LLaMA 3, GPT-4o, and Gemini-1.5-Pro. Each model evaluated one picture description at a time, and provided a binary Yes/No response for each feature. See the complete list of features and prompts in \cref{appendix:full_feature_list}.



To test whether different LLMs can reliably label the 38 binary features, we ran a blind annotation study. Three human annotators independently labeled a subset of the corpus (10 descriptions $\times$ 38 features, totaling $N=380$ values per annotator). Inter-annotator agreement was solid, with Fleiss’ $\kappa = 0.557$. We then applied the Alternative Annotator Test \cite{calderon2025alternative}, a statistical method for evaluating whether LLMs can replace human annotators and for comparing different LLMs. GPT-4o performed best, with a 90\% chance that its answers were as good as or better than those of humans. It outperformed Gemini-1.5-Pro (83\%) and Llama-3.1 (70\%). Notably, GPT-4o passed the test with a conservative threshold ($\varepsilon = 0.1$), limiting the acceptable disagreement between LLM and human annotations, as recommended by \citet{calderon2025alternative}. Given this statistical justification, we used GPT-4o to label all 38 binary features across our 514 descriptions, a total of 19,532 annotated values. \cref{app:data_analysis} details the feature value distribution and internal correlations.

\section{Perception Modeling}
\label{sec:logreg_modeling}


We use stepwise logistic regression, an inherently interpretable model suited for our binary outcomes (Dementia vs. Healthy). The stepwise approach simplifies the model by eliminating weak predictors, and is well-suited for small datasets. We selected logistic regression since, under standard assumptions, particularly when relevant confounders are included, coefficients can roughly approximate causal effects \citep{cinelli2024crash}. In the absence of a formal causal graph, randomized control trials, etc., regression offers a close approximation to causal insights, albeit with caution.

Since our goal is to extract interpretable insights, we fit the model to the full dataset. We evaluate models using McFadden’s $R^2$ \cite{mcfadden1972conditional}, a standard goodness-of-fit metric for logistic regression that quantifies how well the predictors explain the outcome relative to a null model, based on log-likelihood ratio \citep{shtatland2002one, smith2013comparison}. Values between 0.2-0.4 are considered strong and comparable to $R^2$ values of 0.6-0.8 in linear regression \citep{domencich1975urban, louviere2000stated}. To comply with standard NLP practices, we also conduct a complementary analysis of predictive performance and report standard metrics in \cref{app:prediction_evaluation}.

After training we analyze coefficients of features with significant effects \cite{hosmer2013applied}. As an example, the feature \textit{disfluencies} has a coefficient of $\beta = 2.16$ in the LLM perception model. In logistic regression, this means the log-odds of the model assigning a ``Dementia'' label increase by 2.16 when \textit{disfluencies} are present, holding all else constant. Converting this to an odds ratio, the model is approximately 9 times more likely ($e^{2.16} \approx 8.67$) to assign a ``Dementia'' label than a ``Healthy'' one. In contrast, \textit{rich vocabulary} has $\beta = -1.46$ in the LLM perception model, corresponding to an odds ratio of about $e^{-1.46} \approx 0.23$, meaning the model is 
\textit{less likely} to assign a ``Dementia'' label when this feature is present. In general, positive coefficients indicate a shift toward the ``Dementia'' label (1) and negative toward ``Healthy'' (0), with larger values reflecting greater impact.

\section{Results}
\label{sec:results}

\subsection{Modeling Perceptions and Diagnosis}

\begin{figure*}[t]
\label{fig:main_graph}
    \centering

    \begin{minipage}[t]{0.65\textwidth}
        \vspace{0pt}
        \centering
        \includegraphics[width=\linewidth]{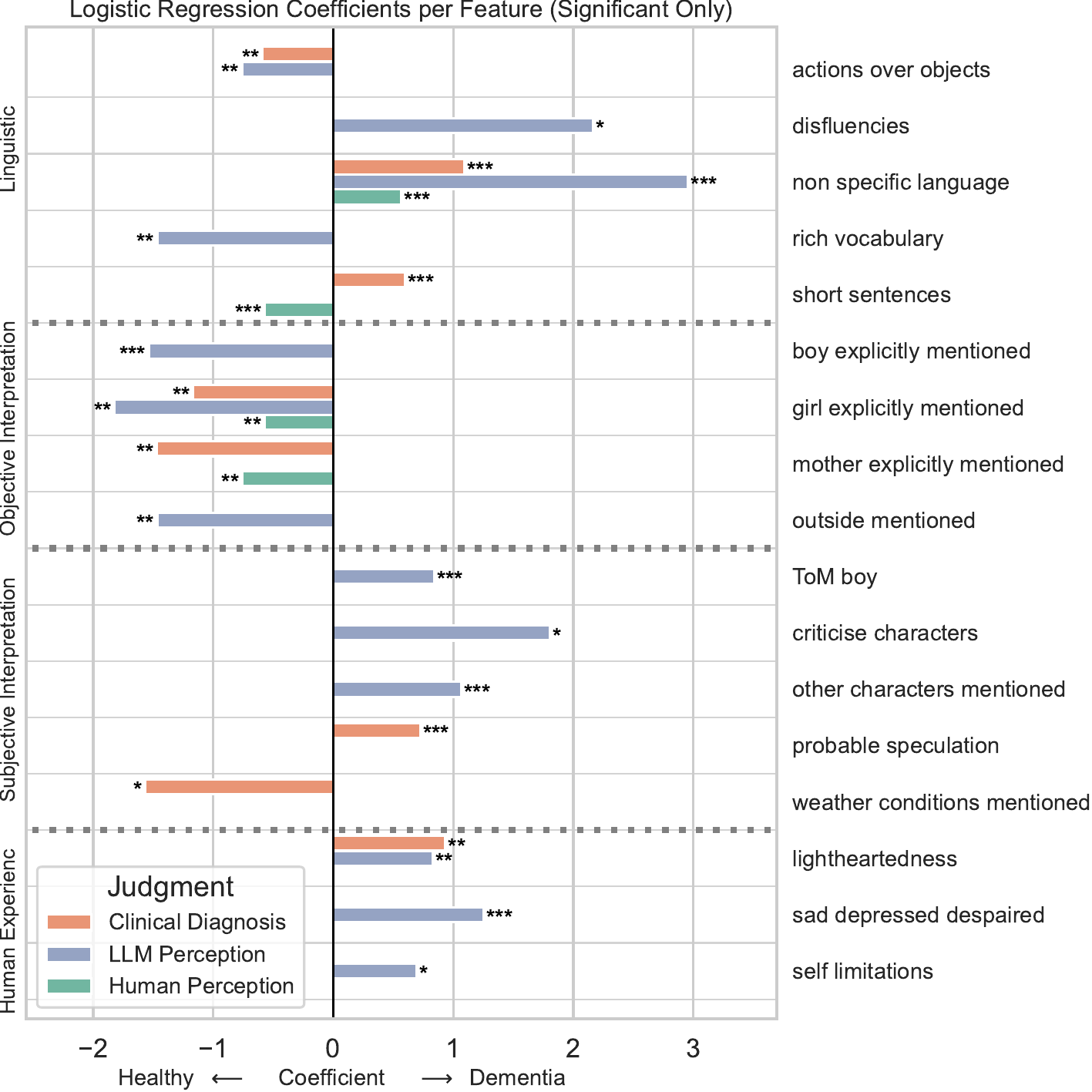}
    \end{minipage}
    \hfill
    \begin{minipage}[t]{0.32\textwidth}
        \vspace{0pt}
        \centering
        \includegraphics[width=\linewidth]{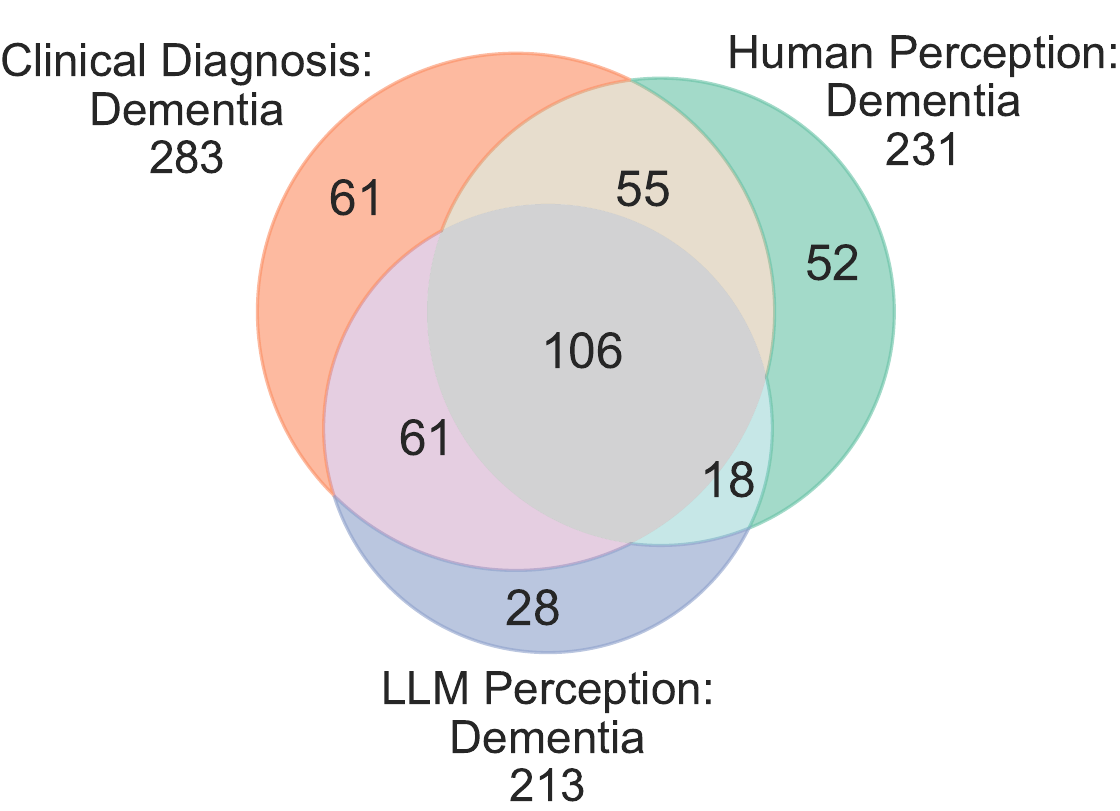}

        \vspace{1em}

        \includegraphics[width=0.8\linewidth]{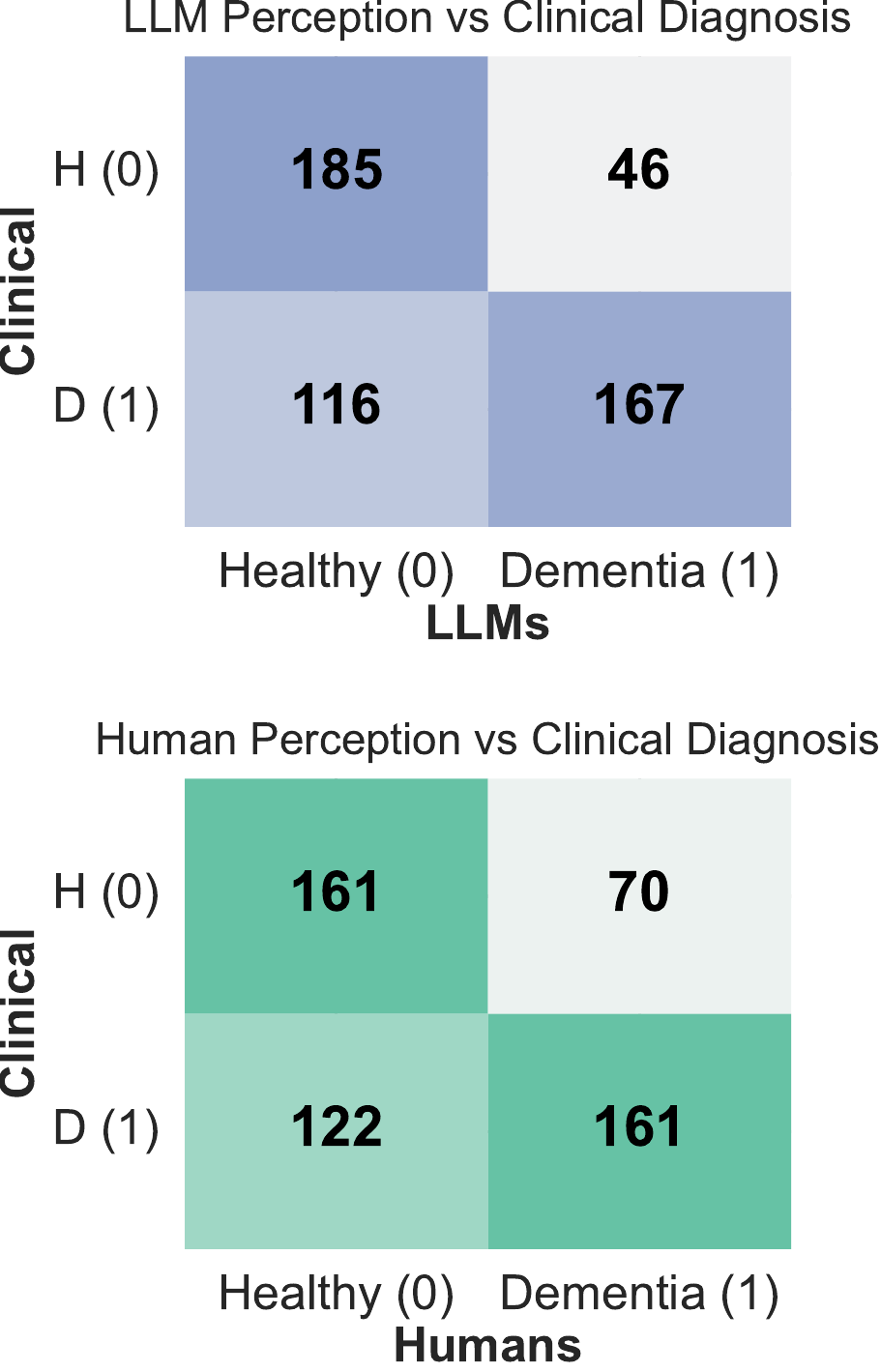}
    \end{minipage}

    \caption{
    Main results from the logistic regression coefficient analysis, and perception disagreements.
\textbf{Left:} Statistically significant features associated with clinical diagnosis, LLM perception, and human perception. Colors indicate the source of judgment; bar direction reflects the sign of the logistic regression coefficient (right = dementia, left = healthy). Dotted lines separate feature categories (Linguistic, Objective Interpretation, etc.). Significance levels: * $^{}p < 0.05$; ** $^{}p < 0.01$; *** $^{}p < 0.001$.
\textbf{Top Right:} Overlap between clinically diagnosed cases and those perceived as dementia by humans and/or LLMs. Of 283 diagnosed, 98 (green, teal, and purple) were missed by humans, LLMs, or both. \textbf{Bottom Right:} Confusion matrices showing alignment between perceptions and diagnosis.
    }
    \label{fig:main_triplet}
\end{figure*}

\cref{fig:main_triplet} illustrates the coefficient values of statistically significant features associated with human perception, LLM judgments, and clinical diagnosis. For all coefficients and corresponding significance values, see \cref{tab:coefficients}, \cref{appendix:full_feature_list}.

McFadden’s $R^ 2$ values indicate a good model fit for clinical diagnosis (0.209) and a very strong fit for LLM dementia perception (0.527), suggesting that our model and features captures a reliable underlying signal. Human perception, however, was harder to model, with McFadden’s $R^2$ = 0.058.

Only a small set of straightforward features are significantly associated with human perception: \textit{non-specific language, short sentences, girl explicitly mentioned}, and \textit{mother explicitly mentioned}, all significantly associated with clinical diagnosis as well. Additional features significantly linked to clinical diagnosis include \textit{actions over objects, other characters mentioned}, and \textit{weather conditions mentioned}. Interestingly, while clinicians associate \textit{short sentences} with dementia, non-experts interpret them as a sign of cognitive health.

Coefficient analysis reveals that all three judgment types are significantly associated with linguistic features such as \textit{non-specific language}, as well as objective interpretations (e.g., whether the boy, girl, or mother is explicitly mentioned). LLMs rely on a broader range of features and categories than clinical diagnoses, showing greater sensitivity to subjective interpretation cues, such as using Theory of Mind (describing others' emotions, intentions, or thoughts) or referring to characters not present in the picture. LLMs also place greater emphasis on emotional expression (\textit{lightheartedness, self-limitations, sad-depressed-despaired}).





\subsection{Diving Into Misperceptions}


We begin with a data analysis, examining how human and LLM judgments align with clinical diagnoses. \cref{fig:main_triplet} presents the corresponding confusion matrices. It also includes a Venn diagram showing the overlap between clinically diagnosed cases and those perceived as dementia by humans and LLMs. Among the 283 clinically diagnosed dementia cases, humans correctly identified 57\%. LLMs, though more conservative in assigning the dementia label, matched 60\%. Some misalignment is expected, as our non-experts rely solely on picture descriptions, while clinical diagnoses draw on a broader range of signals. Errors by both humans and LLMs followed a similar pattern: for LLMs, 70\% of errors were false negatives (i.e., missing clinically diagnosed cases), while 30\% were false positives. Humans showed a similar trend, with a 65-35 ratio, suggesting that both groups have blind spots and room for improvement. 

Next, we examine two key subsets: cases where humans disagreed with both LLMs and clinicians ($n = 121$), e.g., perceiving dementia when both others judged healthy, and cases where LLMs disagreed with both humans and clinicians ($n = 88$). In this analysis, we go beyond comparing perceptions and diagnoses. Instead, we ask: what cues are so prominent that both clinicians and LLMs capture them, but humans miss? And which cues do clinicians and humans detect, but LLMs overlook? We then train a stepwise logistic regression model on each subset to predict human and LLM misperceptions. As before, positive coefficients indicate a shift toward the ``Dementia'' label, and negative toward ``Healthy''-- but here, they reflect \textit{false dementia} and \textit{false healthy}, respectively.

The human misperception model shows a strong fit (McFadden’s $R^2 = 0.6$) and reveals two systematic patterns in human misjudgment: \textbf{(1) Features used differently in misperceived texts:} \textit{non-specific language} received a positive coefficient in the full dataset of 514 samples, indicating that, in general, humans associate non-specific language with dementia. However, in the smaller misperception subset, this feature received a \textit{negative coefficient}, indicating a shift toward the ``Healthy'' label (a \textit{false healthy} judgment, in this context). This suggests that human reliance on this cue is inconsistent-typically treated as a sign of dementia but sometimes overlooked or misinterpreted.

\textbf{(2) Features emerging only in misperceptions:} Some features are not significant in the overall human perception model but become significant in misperception cases. Namely, \textit{rich vocabulary, actions over objects, outside mentioned}, and \textit{boy mentioned} all show a shift toward false healthy judgments, i.e., mistakenly labeling clinically diagnosed dementia cases as healthy. \textit{Lightheartedness}, on the other hand, becomes significant in the misperception model with a shift toward false dementia, suggesting that humans may sometimes associate this cue with cognitive decline. This indicates that certain features, while not consistently relied upon in the general model, may exert misleading influence in specific cases of misjudgment.

The LLM disagreement analysis revealed a quasi-perfect separation: whenever features such as \textit{hesitation}, \textit{reiterating idea}, \textit{grammatical inaccuracies}, or \textit{disfluencies} received a value of `0' (i.e., were not present in the text) the LLM majority vote always labeled the text as ``Healthy''. Other features, namely \textit{sadness, many objects mentioned, mother mentioned, other characters mentioned} were sparse in the disagreement set-- only seven texts expressed sadness, five mentioned other characters, etc. This suggests LLM errors may also stem from data sparsity.




\subsection{Self-reported Human Rationale}


At the end of the task, without having seen our predefined features, annotators answered an open-ended question: Did you notice any patterns in the text that helped with your judgment? These retrospective reports, provided by 18 of 27 participants, may not fully capture real-time reasoning, yet, they offer valuable insights into the behaviors that annotators noticed and said they relied on.

Upon manual review, we found that 65\% of the self-reported cues closely align with our predefined feature set, an encouraging result that suggests our features are naturally noticed by humans. The most frequently cited were \textit{reiterating ideas}, \textit{disfluencies}, and \textit{improbable interpretation}, mentioned by 10, 7, and 5 annotators, respectively. Notably, all four cues revealed as significant for human perception in our coefficient analysis (\cref{fig:main_triplet}) were also self-reported by annotators. A full analysis, including details of newly emergent features from participant responses, is provided in \cref{app:human_rationale}.

\section{Discussion and Insights}
\label{sec:discussion}


\subsection{Human Perception of Dementia}

Modeling human perception proved particularly challenging. The low McFadden’s $R^2$ observed for the human perception model likely reflects not only model limitations but also the inherently noisy and inconsistent nature of human judgment. This is evidenced by low inter-annotator agreement and conflicting annotator rationales, both in the features they cited and the direction of their presumed influence. Notably, these self-reported rationales did not align with the statistically significant features identified by the model, suggesting that people may not rely on what they think they do.

Our model shows that humans often rely on a narrow set of simplistic cues, whereas clinical diagnoses draw on a broader range of signals, including sentiment-related cues, which even LLMs managed to capture. Moreover, humans' misinterpretation of cues-- associating \textit{short sentences} with descriptions from healthy participants, unlike clinical diagnosis-- underscores the importance of education about a broader and more nuanced set of linguistic indicators of dementia.





\subsection{LLMs' Perception}


Unlike humans, LLMs rely on a surprisingly wide set of features, spanning four of our five categories. This use of varied signals may reflect the extensive prior knowledge embedded in their training data, for instance, ``learning'' that Theory of Mind is linked to certain types of dementia \cite{bora2015theory}. This reinforces the importance of rich background knowledge and training to help non-experts and clinicians attend to a broader range of cues.

Both humans and LLMs show a tendency toward false negatives, misjudging dementia cases as healthy. However, in the case of LLMs, a clear pattern emerges: when no linguistic difficulties are present, the model is extremely prone to assigning a ``healthy'' label. Thus, LLMs may be quick to judge a speaker as cognitively healthy if no linguistic dysfunction is apparent, potentially leading to missed signs of early cognitive impairment--an important limitation to recognize and study.


\section{Conclusions}
This study explores how non-expert humans and LLMs perceive dementia in transcribed picture descriptions, and how their perceptions align with clinical diagnoses. We present an inherently interpretable method using high-level, expert-guided features annotated by GPT-4o, followed by logistic regression and coefficient analysis. Ultimately, our work highlights: (1) the inherent difficulty of this task, both in perceiving dementia through text alone and in modeling such a complex phenomenon; (2) the importance of educating both humans and LLMs to recognize a broader range of linguistic signals to improve early detection; and (3) the value of interpretable NLP for advancing research and care across all dementia stakeholders.

\section{Limitations}

Our study makes deliberate modeling and methodological choices to enable a focused analysis of how dementia is perceived. We focus on binary features rather than continuous or categorical ones. While extracting quantitative features (for example, number of verbs) from Pitt corpus' transcripts is a classic and well-studied approach, we pursue a more human-oriented framework that is easier to interpret and is readily translatable into actionable insights. We also experimented with ordinal scales (e.g., ``rate disfluency from 1 to 5''), but during initial evaluations they proved less reliable and consistent, and more difficult to align between human and LLM annotations. Future work could build on our approach and incorporate quantitative metrics such as idea density or lexical diversity to uncover additional signals.

Furthermore, extracting features using LLMs introduces known limitations, such as prompt sensitivity and inconsistency. They may especially be true when extracting features reflecting human intuition, which may not be straightforward for LLMs. While we addressed this by conducting human evaluation and using a statistical test \citep{calderon2025alternative}, future work could expand this approach by testing larger annotator groups and a wider range of language models.

We choose logistic regression for its interpretability, simplicity, and suitability for binary classification tasks such as ours. Its coefficients provide clear, directionally meaningful insights into how individual linguistic features relate to perception labels, aligning with our goal of explainability. However, logistic regression assumes linear relationships and may struggle to capture complex interactions or non-linear patterns present in language data. Additionally, its performance may be limited compared to larger or more expressive models, making it important to balance interpretability with predictive power in future work.

Additionally, we argued that people should consider not only linguistic features and objective interpretations, but also sentiment and subjective interpretations when assessing cognitive health. However, apart from relying on a broader set of cues, another dimension can shape human perception of dementia: long-term personal relationships. In our study, judgments were based on a single written description by an unfamiliar person. In real-life settings, however, impressions are likely to be formed through repeated personal interactions. We assume that participants’ decisions were guided by internalized notions of what a ``typical'' healthy individual sounds like, shaped by general assumptions, societal stereotypes, or portrayals in popular media. This would likely differ if participants were evaluating someone they knew well, drawing on a sense of the individual’s baseline behavior and personality over time.

We use textual picture descriptions only, which do not encompass other modalities that are known to play an important role in discourse (such as motor abilities). While some features (e.g., \textit{disfluencies}) may indirectly reflect physical difficulties in speech production, these are primarily expressed through vocal modalities that are absent from text-based data. In addition, non-verbal gestures (e.g., facial expressions, gaze direction) and prosodic features (e.g., intonation, pitch, speech rate) hold valuable information about cognitive and emotional functioning. These signals are integral to how people perceive and interpret communication in everyday interactions. Future work could address this limitation by incorporating multi-modal inputs, such as audio and video, to capture additional components that shape human perception of dementia in real-world settings.

Finally, our human perception annotations were collected from a contained demographic group. While not representative of the entire population, this group reflects a meaningful demographic: individuals at an age when close relatives may begin to show early signs of cognitive decline. We also acknowledge that while the annotators were proficient in English, based on self-reports and their academic backgrounds, they were not native speakers, which may have influenced their perceptions. In future work, we will involve more diverse populations, including those with varying familiarity with dementia (no exposure, caregivers, clinicians) and broader ranges of age, education, and languages.



\section{Ethical Considerations}

\paragraph{Privacy.} Data confidentiality is a major concern when dealing with clinical data. The dataset used in this study, the Pitt corpus from DementiaBank, is available for research purposes and has already been annonymized. We used only the de-identified transcriptions of the Cookie Theft task, without any accompanying metadata to elicit perceptions. As a result, all data processed in this work was entirely free of personally identifiable information (PII) and handled in accordance with ethical standards.

\paragraph{LLM bias.} LLMs are increasingly used in healthcare research, yet are known to inherit biases from their training data. These biases often reflect societal, linguistic, and cultural norms rather than clinically grounded principles \cite{busch2025current}. They may stem from data imbalance or reliance on proxy variables \cite{chen2018my, obermeyer2019dissecting}. In our work, we explicitly frame model outputs as perceptions to underscore this limitation: LLM outputs do not reflect clinical truth but rather represent patterns the models are able to extract from language data. 

\paragraph{Lack of professional support when relying on observations by non-experts.} When relying on non-experts, particularly close family members or friends, ethical concerns arise regarding their role and the potential impact on relationships. While increasing public awareness of early indicators is valuable, non-experts are not equipped (nor should they be expected) to deliver a diagnosis or communicate life-changing information. Suspicions raised by loved ones can unintentionally undermine trust, introduce stigma, or lead to unjustified restrictions on the autonomy of the person in question. Moreover, such situations can place a significant emotional burden on both parties, as the mere suspicion of a serious and threatening condition like dementia may evoke fear, helplessness, and anxiety. For the person suspected of cognitive decline, being perceived as impaired by someone close can be deeply distressing and lead to feelings of isolation. On the other hand, cases of false negatives, i.e., where early signs are overlooked and diagnosis is delayed,  may result in guilt or self-blame for not having acted sooner. These emotional and relational consequences highlight the need for professional involvement.

The responsibility for diagnosing dementia and delivering difficult news should lie with trained clinicians, who are equipped to do so with sensitivity and to offer appropriate support and care planning. While non-experts may play a valuable role in prompting medical evaluation when they observe concerning changes, they should not carry the burden of diagnostic responsibility. Crucially, failing to notice subtle early symptoms should not be regarded as a personal failure. Dementia is a complex and gradually evolving condition, and even experienced professionals can find its early detection challenging.

\bibliography{anthology,custom}
\bibliographystyle{acl_natbib}

\appendix

\section{Full Feature List and Prompts}
\label{appendix:full_feature_list}

{\scriptsize
\begin{table*}[]
\resizebox{\textwidth}{!}{%
\begin{tabular}{|l|l|p{8cm}|l|}
\hline
\rowcolor[HTML]{EFEFEF} 
\textbf{Category}                                                                                                        & \textbf{Feature Name}                                                 & \textbf{Sources}                                                                                                                                                                                                                                                                                                                                           & \textbf{Example}                                                                                                 \\ \hline
\cellcolor[HTML]{CDD6FF}                                                                                                 & \begin{tabular}[c]{@{}l@{}}Circumlocution\\ (Wordiness)\end{tabular} & \citet{nicholasSystemQuantifyingInformativeness1993, kaveSeverityAlzheimersDisease2018, choAutomatedAnalysisLexical2021}                                                                                                                                                                                                                 & \begin{tabular}[c]{@{}l@{}}``a stool which is about and\\ he he is getting a cookie''\end{tabular}                 \\ \cline{2-4} 
\cellcolor[HTML]{CDD6FF}                                                                                                 & Grammatical Inaccuracies                                              & \citet{croisileComparativeStudyOral1996, fraserLinguisticFeaturesIdentify2016}                                                                                                                                                                                                                                                           & ``they be stealing''                                                                                               \\ \cline{2-4} 
\cellcolor[HTML]{CDD6FF}                                                                                                 & Introduction                                                          & \citet{ortiz-perezDeepLearningBasedMultimodal2023}                                                                                                                                                                                                                                                                                       & ``This is a family scene. A boy...''                                                                               \\ \cline{2-4} 
\cellcolor[HTML]{CDD6FF}                                                                                                 & Naming Characters                                                     & \citet{kempler1994language, kempler2008language}                                                                                                                                                                                                                                                                                         & ``Johnny here is...''                                                                                              \\ \cline{2-4} 
\cellcolor[HTML]{CDD6FF}                                                                                                 & Non-specific language                                                 & \citet{nicholasSystemQuantifyingInformativeness1993, kaveSeverityAlzheimersDisease2018, cummingsDescribingCookieTheft2019, choAutomatedAnalysisLexical2021}                                                                                                         & ``The thing there..''                                                                                              \\ \cline{2-4} 
\cellcolor[HTML]{CDD6FF}                                                                                                 & Rich vocabulary                                                       & \citet{fraserLinguisticFeaturesIdentify2016, kaveSeverityAlzheimersDisease2018, choAutomatedAnalysisLexical2021, williamsLexicalsemanticPropertiesVerbs2023}                                                                                                                                                                             & \begin{tabular}[c]{@{}l@{}}``She is probably daydreaming. \\ The faucet...''\end{tabular}                          \\ \cline{2-4} 
\cellcolor[HTML]{CDD6FF}                                                                                                 & Short sentences                                                       & \citet{mueller2016connected, fraserLinguisticFeaturesIdentify2016}, \citet{kave2003morphology, forbes2005detecting}                                                                                                                                                                       & ``Stealing cookies. Falling stool.''                                                                               \\ \cline{2-4} 
\cellcolor[HTML]{CDD6FF}                                                                                                 & Starts with interjection                                              & \citet{karlekarDetectingLinguisticCharacteristics2018, ortiz-perezDeepLearningBasedMultimodal2023}                                                                                                                                                                                                                                       & ``Well. Mother is..''                                                                                              \\ \cline{2-4} 
\cellcolor[HTML]{CDD6FF}                                                                                                 & Disfluencies                                                          & \citet{nicholasSystemQuantifyingInformativeness1993, szatloczkiSpeakingAlzheimersDisease2015, muellerDeclinesConnectedLanguage2018, kumarMLBasedAnalysisIdentify2021}& ``She is uh, uh, umm..''                                                                                           \\ \cline{2-4} 
\cellcolor[HTML]{CDD6FF}                                                                                                 & Self corrections                                                      & \citet{rudziczAutomaticallyIdentifyingTroubleindicating2014, muellerDeclinesConnectedLanguage2018}                                                                                                                                                                                                                                       & \begin{tabular}[c]{@{}l@{}}``Mother's washing, uh, \\ drying dishes''\end{tabular}                                 \\ \cline{2-4} 
\multirow{-11}{*}{\cellcolor[HTML]{CDD6FF}\textbf{Linguistic}}                                                           & Actions over objects                                                  & \citet{williamsLexicalsemanticPropertiesVerbs2023, kaveSeverityAlzheimersDisease2018}                                                                                                                                                                                                                                                    & \begin{tabular}[c]{@{}l@{}}``She's Washing, they're stealing''\\ (vs. ``I see curtains, shoes'')\end{tabular}        \\ \hline
\cellcolor[HTML]{E8CDFF}                                                                                                 & Boy explicitly mentioned                                              & \citet{croisileComparativeStudyOral1996, nicholasSystemQuantifyingInformativeness1993, cummingsDescribingCookieTheft2019}                                                                                                                                                                                                                & ``A young lad here''                                                                                               \\ \cline{2-4} 
\cellcolor[HTML]{E8CDFF}                                                                                                 & Girl explicitly mentioned                                             & \citet{croisileComparativeStudyOral1996, nicholasSystemQuantifyingInformativeness1993, cummingsDescribingCookieTheft2019}                                                                                                                                                                                                                & ``Sister is laughing''                                                                                             \\ \cline{2-4} 
\cellcolor[HTML]{E8CDFF}                                                                                                 & Kitchenware attention                                                 & \citet{croisileComparativeStudyOral1996, nicholasSystemQuantifyingInformativeness1993, cummingsDescribingCookieTheft2019}                                                                                                                                                                                                                & ``Two cups and a plate''                                                                                           \\ \cline{2-4} 
\cellcolor[HTML]{E8CDFF}                                                                                                 & Mother explicitly mentioned                                           & \citet{croisileComparativeStudyOral1996, nicholasSystemQuantifyingInformativeness1993, cummingsDescribingCookieTheft2019}                                                                                                                                                                                                                & ``Mother is...''                                                                                                   \\ \cline{2-4} 
\multirow{-5}{*}{\cellcolor[HTML]{E8CDFF}\textbf{\begin{tabular}[c]{@{}l@{}}Objective \\ Interpretation\end{tabular}}}   & Outside mentioned                                                     & \citet{croisileComparativeStudyOral1996, nicholasSystemQuantifyingInformativeness1993, cummingsDescribingCookieTheft2019}                                                                                                                                                                                                                & ``The garage is...''                                                                                               \\ \hline
\cellcolor[HTML]{FFE1CD}                                                                                                 & Assumes sound                                                         & \citet{yorkstonAnalysisConnectedSpeech1980}                                                                                                                                                                                                                                                                                              & ``Sister is saying...''                                                                                            \\ \cline{2-4} 
\cellcolor[HTML]{FFE1CD}                                                                                                 & Cause and effect                                                      & \citet{croisileComparativeStudyOral1996, cummingsDescribingCookieTheft2019}                                                                                                                                                                                                                                                              & ``Boy is about to fall''                                                                                           \\ \cline{2-4} 
\cellcolor[HTML]{FFE1CD}                                                                                                 & Criticise characters                                                  & \citet{yorkstonAnalysisConnectedSpeech1980}                                                                                                                                                                                                                                                                                              & ``Mother is doing a bad job''                                                                                      \\ \cline{2-4} 
\cellcolor[HTML]{FFE1CD}                                                                                                 & Empathy                                                               & \citet{chowMedialTemporalLobe2023, demichelisEmpathyTheoryMind2020}                                                                                                                                                                                                                                                                      & ``Poor boy is about to fall''                                                                                      \\ \cline{2-4} 
\cellcolor[HTML]{FFE1CD}                                                                                                 & Improbable interpretation                                             & \citet{nicholasSystemQuantifyingInformativeness1993}                                                                                                                                                                                                                                                                                     & ''Pineapple jar''                                                                                                  \\ \cline{2-4} 
\cellcolor[HTML]{FFE1CD}                                                                                                 & Mentioning many objects                                               & \citet{nicholasSystemQuantifyingInformativeness1993, williamsLexicalsemanticPropertiesVerbs2023}                                                                                                                                                                                                                                         & ``Shoes, dress, cupboard handles...''                                                                              \\ \cline{2-4} 
\cellcolor[HTML]{FFE1CD}                                                                                                 & Other characters mentioned                                            & \citet{nicholasSystemQuantifyingInformativeness1993}                                                                                                                                                                                                                                                                                                                                                         & ``Baby is crying in the other room''                                                                               \\ \cline{2-4} 
\cellcolor[HTML]{FFE1CD}                                                                                                 & Probable speculation                                                  & \citet{nicholasSystemQuantifyingInformativeness1993}                                                                                                                                                                                                                                                                                     & ``Sink is probably clogged''                                                                                       \\ \cline{2-4} 
\cellcolor[HTML]{FFE1CD}                                                                                                 & Theory of Mind- Boy                                                   & \citet{cummingsDescribingCookieTheft2019, demichelisEmpathyTheoryMind2020, zegarra-valdiviaCognitiveEmotionalTheory2023}                                                                                                                                                                               & \begin{tabular}[c]{@{}l@{}}``Boy is handing a cookie \\ to his sister''\end{tabular}                               \\ \cline{2-4} 
\cellcolor[HTML]{FFE1CD}                                                                                                 & Theory of Mind- Girl                                                  & \citet{cummingsDescribingCookieTheft2019, demichelisEmpathyTheoryMind2020, zegarra-valdiviaCognitiveEmotionalTheory2023}                                                                                                                                                                               & ``Sister is laughing...''                                                                                          \\ \cline{2-4} 
\cellcolor[HTML]{FFE1CD}                                                                                                 & Theory of Mind- Mother                                                & \citet{cummingsDescribingCookieTheft2019, demichelisEmpathyTheoryMind2020, zegarra-valdiviaCognitiveEmotionalTheory2023}                                                                                                                                                                               & \begin{tabular}[c]{@{}l@{}}``Mother is thinking about \\ something...''\end{tabular}                               \\ \cline{2-4} 
\multirow{-12}{*}{\cellcolor[HTML]{FFE1CD}\textbf{\begin{tabular}[c]{@{}l@{}}Subjective \\ Interpretation\end{tabular}}} & Weather conditions mentioned                                          & \citet{nicholasSystemQuantifyingInformativeness1993}                                                                                                                                                                                                                                                                                     & ``It's a sunny day''                                                                                               \\ \hline
\cellcolor[HTML]{E0F5DE}                                                                                                 & Checking Previously Said                                              & \citet{cipriani2013repetitive}                                                                                                                                                                                                                                                                                                           & ``Did I already say that?''                                                                                        \\ \cline{2-4} 
\cellcolor[HTML]{E0F5DE}                                                                                                 & Continuing after saying done                                          & \citet{baylis2004visual}                                                                                                                                                                                                                                                                                                                 & ``That's all I can see. They boy is...''                                                                           \\ \cline{2-4} 
\cellcolor[HTML]{E0F5DE}                                                                                                 & Hesitation                                                            & \citet{rudziczAutomaticallyIdentifyingTroubleindicating2014, ortiz-perezDeepLearningBasedMultimodal2023}                                                                                                                                                                                                                                 & \begin{tabular}[c]{@{}l@{}}``I'm not sure, maybe stealing \\ cookies''\end{tabular}                                \\ \cline{2-4} 
\cellcolor[HTML]{E0F5DE}                                                                                                 & Irritability                                                          & \citet{cummingsDescribingCookieTheft2019}                                                                                                                                                                                                                                                                                                & ``Can I go now?''                                                                                                  \\ \cline{2-4} 
\cellcolor[HTML]{E0F5DE}                                                                                                 & Lightheartedness                                                      & \citet{cummingsDescribingCookieTheft2019, granitsasAllLaughterNervous2020}                                                                                                                                                                                                                                                               & \begin{tabular}[c]{@{}l@{}}``Wowie! This is a mess [laughs]''\end{tabular}                          \\ \cline{2-4} 
\cellcolor[HTML]{E0F5DE}                                                                                                 & Reiterating ideas                                                     & \citet{croisileComparativeStudyOral1996, stegmannComparisonRemoteInperson2021, kumarMLBasedAnalysisIdentify2021}                                                                                                                                                                                                                         & \begin{tabular}[c]{@{}l@{}}``Boy is stealing. Mother is \\ washing dishes. Boy is taking \\ cookies''\end{tabular} \\ \cline{2-4} 
\cellcolor[HTML]{E0F5DE}                                                                                                 & Sad depressed despaired                                               & \citet{cummingsDescribingCookieTheft2019}                                                                                                                                                                                                                                                                                                & \begin{tabular}[c]{@{}l@{}}``Please stop, I can't take this \\ anymore''\end{tabular}                              \\ \cline{2-4} 
\cellcolor[HTML]{E0F5DE}                                                                                                 & Self limitations                                                      & \citet{rudziczAutomaticallyIdentifyingTroubleindicating2014, nicholasSystemQuantifyingInformativeness1993}                                                                                                                                                                                                                               & ``I don't see anything else.''                                                                                     \\ \cline{2-4} 
\multirow{-9}{*}{\cellcolor[HTML]{E0F5DE}\textbf{\begin{tabular}[c]{@{}l@{}}Human \\ Experience\end{tabular}}}           & Vision difficulties                                                   & \citet{lawrence2009out}                                                                                                                                                                                                                                                                                                                  & ``Let me put on my glasses''                                                                                       \\ \hline
\cellcolor[HTML]{F7D5D5}\textbf{\begin{tabular}[c]{@{}l@{}}Interview\\ Setting\end{tabular}}                             & Clarification required                                                & \citet{rudziczAutomaticallyIdentifyingTroubleindicating2014, karlekarDetectingLinguisticCharacteristics2018}                                                                                                                                                                                                                             & ``Did you want only actions?''                                                                                     \\ \hline
\end{tabular}%
}
\caption{Full feature list, literature sources and examples.}
\label{tab:full_features}
\end{table*}
}

Following is the complete list of 38 binary features we defined, divided per category. For each feature, we present its relevant prompt, provided to GPT-4o. 

\subsection{Linguistic Features}
These are features representing the speaker's use of language.

\paragraph{Circumlocution:} ``You are analyzing a transcribed description of the Cookie Theft Picture from the Boston Diagnostic Aphasia Examination (BDAE). Your task is to determine whether the speaker shows signs of circumlocution excessive or indirect speech that makes it harder to follow the message.
Answer ``yes'' only if the text contains noticeable wordiness or overly long explanations that make the message less clear or harder to follow. This includes:
* Describing something in a roundabout way without clearly naming it (e.g., ``the thing you use to dry stuff'' instead of ``towel'').
* Over-explaining simple observations (e.g., ``and then I guess what she’s doing is maybe she’s holding something and it looks like it could be'' instead of directly saying ``she’s holding a plate'').
Disregard normal conversational markers like-
* Slight elaboration or descriptive phrasing.
* Minor hesitations or informal language.
Please answer yes or no only, no explanations.

\begin{figure}[t]
    \centering
    \includegraphics[width=1\linewidth]{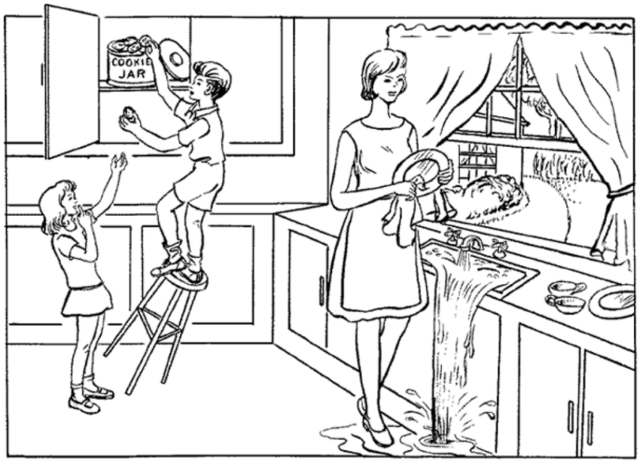}
    \caption{Cookie Theft Picture.}
    \label{fig:cookie}
\end{figure}

\paragraph{Grammatical inaccuracies:} ``Does the text contain noticeable grammatical errors (e.g., ``the boy going'', ``they is washing dishes'', ``the boy does climbing'')? Does not qualify: (1) Incomplete sentences without clear content, (2) Hesitations and disfluencies (e.g., ``the boy went, uh, boy'') unless they result in a grammatical mistake. Please answer yes or no only, no explanations.''

\paragraph{Introduction:} ``Is the first sentence a general introduction to the picture? Qualifies: (1) A statement that sets up the scene before describing specific elements (e.g., ``This is a family scene.'', ``This is a mess.''). (2) A general remark about the picture, even if opinionated or informal (e.g., ``What a mess in this kitchen!''). Does not qualify: Starting with fillers ('uh. the boy is...') or discourse markers ('Well, this here is a cookie jar.'). (2)  Jumping straight into the description without an introduction (e.g., ``Okay, the mother is cooking.'') (3) Making statements about the task ('I'll start now.') (4) asking questions/clarifications directed at the interviewer (e.g., ``Can I start?'', ``You wanted only actions?''). Please answer yes or no only, no explanations.''

\paragraph{Naming characters:} ``Does the speaker call any of the characters by any name? (e.g., ``Johnny here is stealing cookies'') please answer with ``yes'' or ``no'' only, no explanation.''

\paragraph{Non-specific language:} ``You are analyzing a transcribed description of the Cookie Theft Picture from the Boston Diagnostic Aphasia Examination (BDAE). Your task is to determine whether the speaker exhibits a strong reliance on non-specific language--including both non-specific nouns (e.g., excessive pronoun use or vague descriptions) and non-specific verbs (e.g., ''doing something'', ``going over there'')--in a way that suggests difficulty retrieving specific words.
A description should be flagged as ``yes'' only if:
There is a strong and persistent pattern of avoiding specific words, making the description vague or unclear.
Example: ``He's on that thing, and she's doing something over there.''
Example: ``The thing is falling, and she's making the water go.''
The avoidance applies to both nouns and verbs, reducing clarity significantly.
The pattern is excessive, indicating possible word retrieval difficulty rather than normal variation in speech.
Examples (Flag as ``yes''):
``That one is up on the thing, and the other one is doing something.''
``She’s making it go while he’s messing with that over there.''
``They’re all doing things, and it’s happening.''
Examples (Flag as ``no''):
``The boy is standing on the stool, and the mother is washing dishes.'' (Clear noun and verb use.)
``She is drying the dishes while he is reaching for cookies.'' (Some pronoun use but still specific enough.)
Response Format:
Return ``yes'' if the description contains an excessive reliance on non-specific language (both nouns and verbs) to the point that it strongly suggests word retrieval difficulty.
Return ``no'' if pronoun or vague verb use is within normal variation or does not significantly impair clarity.
Please return ``yes'' or ``no'' only, no intermediate calculations or explanations.''

\paragraph{Rich vocabulary:} ``You are analyzing a spoken or written description of the Cookie Theft Picture from the Boston Diagnostic Aphasia Examination (BDAE). Your task is to determine whether the text demonstrates vocabulary richness.
Respond with ``yes'' if the description includes specific and varied content words, such as precise nouns, vivid verbs, coumpund nouns or relatively unique phrases (e.g., ``overflowing'', ``spigot'', ``faucet'', ``summertime'', ``drapes''). These indicate the speaker is using a rich and varied vocabulary beyond basic object naming.
Respond with ``no'' if the description is dominated by vague, repetitive, or basic words, lacks elaboration, or relies heavily on simple naming without further detail. Frequent disfluencies or abandoned thoughts may also suggest limited vocabulary use.
Return ``yes'' or ``no'' only. No explanations.''

\paragraph{Short sentences:} ``Does the speaker primarily use short, independent clauses with minimal subordination or conjunctions? Responses should favor simple sentence structures (e.g., 'The boy is in the cookie jar. The mother spilled the water. The window is open.') over complex or compound sentences with multiple dependent clauses (e.g., 'The young fellow is standing on a stool which is getting ready to fall while he's handing the top of the cookie jar.'). Please answer 'yes' or 'no' only, no explanations.''

\paragraph{Starts with interjection:} ``Does this text start with discourse markers (e.g., 'well', 'like', 'you know', 'I mean', 'okay'), verbal fillers (e.g., 'let me think...', 'the thing is...', 'how do I put this...'), or repetition (e.g., 'so, so...', 'you know, you know...')? Does not qualify: (1) starting with empty speach and fillers (e.g., 'uh', 'umm'), (2) starting with content words like ``cookie jar''. Please answer yes or no only, no explanation.''

\paragraph{Disfluencies:} ``You are analyzing a transcribed description of the Cookie Theft Picture from the Boston Diagnostic Aphasia Examination (BDAE). Your task is to determine whether the text contains a noticeable amount of disfluencies.
These include:
(1) Repeated starts (e.g., ``the girl... uh... the mother'') signaling a change of mind or confusion.
(2) Incomplete or abandoned utterances (e.g., ``the girl went the boy'').
(3) Unusual or excessive fillers and hesitations (e.g., many uses of =``uh'' or ``um'') suggesting trouble forming thoughts rather than conversational rhythm.
Remember that this is transcribed speech. So you should disregard normal spoken features, such as:
(1) Occasional ``uh'' or ``um'' used naturally.
(2) Minor self-corrections or restarts that do not break sentence coherence.
Answer only with ``yes'' or ``no'', with ``yes'' reserved for texts presenting excessive disfluencies.''

\paragraph{Self corrections:} ``Does the text include self-corrections? i.e., the speaker says something, identify they have a mistake, then immediately corrects themselves (e.g., ``a kid uh, getting in the... falling off the stool'', ``the girl, uh, mother''). Hesitations and fillers by themselves (e.g., ``the girl, uh, uh... is climbing'') do not count. Please answer yes or no only, no explanations.''

\paragraph{Actions over objects:} ``Does the speaker focus on actions rather than objects, showing a tendency toward describing movement and processes over static elements (e.g., ``reaching'', ``tipping'', ``putting'' instead of ``cookie jar'', ``stool'', ``apron'', ``shoes''). Please answer yes or no only, no explanations.''

\subsection{Objective Interpretation features}

These features represent whether the speaker refers to elements that are indisputably visible in the picture.

\paragraph{Boy explicitly mentioned:} ``This text describes a picture including a boy in the context of climbing a stool and reaching for a cookie jar. Is this boy explicitly, unmistakable mentioned? It qualifies in both direct terms (e.g., 'boy', 'lad', 'man', 'youngster', 'junior') and indirect terms ('he', 'him', 'Johnny'), provided that the context is clear.  If the speaker describes children in plural (e.g., 'children', 'kids'), it qualifies. Please answer ``yes'' or ``no'' only, no explanations. If you are uncertain if the speaker refers to the boy, please say ``no''.''

\paragraph{Girl explicitly mentioned:} ``This text describes a picture including a girl in the context of standing next to a boy, plotting to steal cookies, putting a finger to her lips, or reaching for the boy. Is this girl explicitly, unmistakably mentioned? It qualifies in both direct terms (e.g., 'girl', 'sister') and indirect terms ('her', 'she'), provided that the context unambiguously refers to the girl in a way that is evident in the picture. If the speaker describes children in plural (e.g., 'children', 'kids'), it qualifies. If you are uncertain whether the speaker refers to the girl, say ``no''. Otherwise, answer ``yes'' only.''

\paragraph{Kitchenware attention:} ``Your task is to determine whether the speaker pays attention to kitchenwear items (e.g., tap, towels, cups, plates) Positive example: ''cups on the counter and dish on the counter'', ``washing this plate''. Mentioning ``dishes'' as a general term, or as part of the mother's (she's drying dishes) does not count, since it does not focus on which dishes. ``cookie jar'' does not qualify as kitchenwear items. Please answer with ``yes'' or ``no'' only, no explanation.''

\paragraph{Mother explicitly mentioned:} ``This text is a description of the Cookie Theft Picture. The picture includes a mother in the context of washing and drying dishes around a sink with spilling water. She is wearing a dress and shoes, handling dishes, her feet are in water, and she is looking outside the window. Is this character explicitly, unmistakably mentioned? Both direct terms (mother, lady, woman, model) and indirect terms (she, her) qualify, given that the context of water, dishes, etc is clear.  Even if the speaker portrays uncertainty (e.g., ``I guess''), explicitly stating ``mother'' still qualifies. Any female characters mentioned in the contexts of dishes or standing in water is the mother. Any female mentioned in context of cookie jar is not the mother so does not qualify.''

\paragraph{Outside mentioned:} ``Does the speaker mention the physical outside surroundings of the house? (i.e. bushes, trees, path, sun) Weather conditions (e.g. wind, summer) do not count. Please answer with ``yes'' or ``no'' only, no explanation.''

\subsection{Subjective Interpretation features}

These features represent whether the speaker presents higher-level interpretations that may or may not be accurate.

\paragraph{Assumes sound:} ``Does the speaker explicitly describe speaking, or any noises being made (such as speaking, shushing, birds chirping, a kettle whistling, or other sounds)? Qualifies: mentions of sounds OR sound-inducing actions (e.g., talking, laughing) produced by characters or objects in the scene. Do not assume a sound is present based on an event (e.g., dropping a plate) unless the speaker explicitly states that a noise occurred. Please answer with ``yes'' or ``no'' only, no explanation.''

\paragraph{Cause and effect:} ``Does the text contain an inference or prediction about something probable that is happening now or is about to happen, based directly on the scene?
Answer ``yes'' if the text includes a logical inference or prediction stemming clearly from the picture (e.g., ``soon there will be a mess'', ``the boy is going to fall'', ``the boy is falling''). The inferred event must be a direct consequence of what is visible in the image.
Answer ``no'' if the statement (1) does not contain inference or prediction, (2) makes assumptions about imagined events unrelated to the immediate scene (e.g., ``the husband will be back from work''), or (3) is too unclear without an obvious cause-and-effect link to the picture.
Respond with ``yes'' or ``no'' only. No explanation.''

\paragraph{Criticise characters}: ``Does the speaker explicitly criticize the characters in the scene? (e.g., ``she is irresponsible'', ``I would spank that girl'', ``they are reckless''). Criticism includes negative judgment or disapproval directed at what the characters are doing, how they are behaving, or the choices they are making (e.g., ``that looks dangerous!''). Does not qualify: general frustration towards the task itself (``this is stupid'', ``can I leave now'' etc.) please answer 'yes' or 'no' only, no explanations.''

\paragraph{Empathy:} ``Does the speaker use words to communicate empathy toward the characters in the picture? Answer 'yes' if the text includes ANY explicit expressions of emotional concern, sympathy, or understanding (e.g., 'poor mama,' 'that must hurt,' 'she looks worried,' or 'the kids are nicely dressed'). Only direct emotional expressions count! Does not count: descriptive statements (e.g., 'the boy is going to fall and hurt himself'). Please answer 'yes' or 'no' only, no explanations.''

\paragraph{Improbable interpretation:} ``Does this text include anything entirely unrelated to the Cookie Theft image? Return ``yes'' if you are confident that the description includes elements that does not correspond to anything that could reasonably be inferred from the scene, like an impossible or completely fabricated item or thing.''

\paragraph{Mentioning many objects:} ``Does the text describe the scene primarily by listing a large number of inanimate objects (e.g., shoes, clothes, dishes, sink, cookie jar, curtains, windows, cupboard, lid) rather than focusing on people and their actions?
Answer ``yes'' if the description is dominated by objects, details, and static elements rather than actions or interactions. Answer ``no'' if the focus is primarily on people and their actions, even if some objects are mentioned.
Respond with ``yes'' or ``no'' only. No explanation.''

\paragraph{Other characters mentioned:} ``Does the text include human characters that are related to the scene but are not the mother (explicitly or implied), girl (explicitly or implied), or boy (explicitly or implied)? A mention qualifies even if the speaker later questions, dismisses, or negates the relationship (e.g., 'This wouldn’t be the grandmother' and 'I don't see a husband' still qualify). Please answer with 'yes' or 'no' only, no explanations.''

\paragraph{Probable speculation:} ``Does the speaker describe something that is not explicitly visible in the picture but is a reasonable interpretation based on the scene?
Answer ``yes'' if the speaker makes an assumption that cannot be directly proven or disproven but makes sense in context (e.g., ``it's a nice day outside'' because the window is open and people are in short sleeves, or ``the husband is at work'' in a family setting). Answer ``no'' if: (1) the text is strictly based on what is visible in the image, without adding interpretations beyond the scene, (2) the text is completely unclear, (3) the inference is completely nonsensical. Respond with ``yes'' or ``no'' only. No explanation.''

\paragraph{ToM boy:} ``Does the speaker use Theory of Mind when describing the boy? Includes: Assuming his mood or intentions (e.g., ``poor boy'', ``he's generous''), Subjective interpretations of his actions (e.g., assuming he is stealing, meaning it's forbidden), Inferring readiness or preparation to act (e.g., ``he's about to give her a cookie''), References to the boy explicitly or by other terms ('Johnny', 'man', 'kid', 'him', 'brother', etc.), Plural references that apply ToM to both children (``they are stealing'')
Does not qualify: Objective, indisputable descriptions of the boy's actions (e.g., ``taking/reaching for a cookie'')
Speculations about future occurrences (e.g., ``he is about to fall'') that do not involve intent or mental states.'' 

\paragraph{ToM girl:} ``Does the speaker use Theory of Mind (ToM) when describing the girl? This includes projecting her mood, mental state, intentions, or subjective interpretations of her actions (e.g., assuming she is happy, sneaky, has a certain motivation, is saying something). It also qualifies if ToM is applied to both kids in plural (``they are stealing''). Does not qualify: ToM applied to the mother. If 'she' or 'her' is used, determine from context whether it refers to the mother or the girl. please answer with ``yes'' or ``no'' only, no explanation.''

\paragraph{ToM mother:} ``Does the speaker use Theory of Mind (ToM) when describing the mother? This includes projecting her mood, mental state, intentions, or subjective interpretations of her actions (e.g., assuming she is absent-minded, inattentive, unconcerned about her children, or responsible for the mess). ToM includes descriptions implying the mother is thinking about or focusing on something specific (e.g., ``she is looking outside'', ``she is unaware of the water'') rather than purely describing physical actions indisputibly portrayed in the picture. ToM applied to the girl does not count. If 'she' or 'her' is used, determine from context whether it refers to the mother or the girl.''

\paragraph{Weather conditions mentioned:} ``Does the speaker mention any weather conditions or seasons of the year? (i.e. summer, windy, a hot day, etc.) Please answer with ``yes'' or ``no'' only, no explanation.''

\subsection{Human Experience:}

These features represent whether the speaker expresses their own state or emotions.

\paragraph{Checking previously said:} ``Does the speaker explicitly check if they've already mentioned something in their description? Qualify: direct questions or statements such as ``Did I already say...?'', that indicate they are actively verifying whether they have said something before. Does not qualify: Self-corrections or rewording a previous statement. Does not qualify: reiterating ideas (``I already told you that...''). Only checking whether something was already said. Please answer with ``yes'' or ``no'' only, no explanations.''

\paragraph{Continuing after saying done:} ``A contradiction is when someone claims they are done describing the picture (e.g., ``that's all'', ``I'm done'', ``that's all I can see'') but afterwards continues to provides more knowledge about the picture (e.g. continues talking about the jar, house, etc). If you note such a contradiction, return ``yes''. Otherwose, return ``no''. Return only ``yes'' or ``no'', no explanations or intermediate calculations.''

\paragraph{Hesitation:} ``Does the speaker explicitly, verbally express hesitation in their own description?
Qualifies: (1) Direct phrases such as ``I'm not sure'', ``I don't know.'' (2) Fragmented, self-questioning speech (e.g., ``uh... which one?'' ``this... thing?''). (3) Self-directed questioning about how to describe something (e.g., ``now what would I say about them?'' ``how do I put this?'').
Does not qualify: (1) Speculative statements (e.g., ``could be'', ``he will probably fall''). (2) Words expressing a certain level of certainty like ``probably'', ``I guess'', ``I think'' (3) Statements of inability (e.g., ``I can't see anything else'', ``I can't make out any action''). (4) Questions to the interviewer (e.g., ``is that it?'' ``did I do it right?''). Please answer ``yes'' or ``no'' only, no explanation.''

\paragraph{Irritability:} ``Does the speaker show indisputable irritation? please answer with ``yes'' or ``no'' only, no explanation.''

\paragraph{Lightheartedness:} ``Does the speaker explicitly convey humor or lightheartedness through their tone or choice of words? This includes the speaker laughing, joking, playful expressions (e.g., ``gee whiz'', ``oh boy'', ``golly'', ``oh goody''), or an amused tone. The description of a chaotic or absurd scene does not qualify unless the speaker’s words or manner explicitly suggest they find it humorous or lighthearted. please answer with ``yes'' or ``no'' only, no explanation.''

\paragraph{Reiterating idea:} ``Does the speaker repeat the same idea multiple times within the text? i.e., reintroducing a previously mentioned element after discussing other details (e.g., idea A, idea B, idea C, idea A) please answer with ``yes'' or ``no'' only, no explanation.''

\paragraph{Sad depressed despaired:} ``Does the speaker sound depressed, sad or despaired? please answer with ``yes'' or ``no'' only, no explanation.''

\paragraph{Self limitations:} ``Does the speaker explicitly express a personal limitation in their ability to complete the task, like self-deprecation? This includes statements indicating that they cannot see, understand, or describe something (e.g., ``I can't see anything else'', ``I don't know what that is'', ``I can't do this'', ``I give up.''). Do not qualify: (1) Expressions of frustration without implying a personal limitation (e.g., ``this is dumb'', ``this is annoying'') (2) questions or requests to stop (e.g., ``Can I stop now?'' ``Do I have to keep going?'') (3) general finishing statements like ``that's all'', ``I guess that's it''. please answer with ``yes'' or ``no'' only, no explanation).''

\paragraph{Vision difficulties:} ``Can it be understood from the text that the speaker has vision impairment? Either explicitly (``where are my glasses'') or implicitly (``what does it say there?'') please answer with ``yes'' or ``no'' only, no explanation.''

\subsection{Interview Context Features}

This feature represents whether the text includes references to the interview situation itself.

\paragraph{Clarification required:} ``Please create a list of sentences from this text that contain questions or exchange of words with someone. For each, rank 1-5 if it makes sense that this was said to the interviewer, or spoken to oneself. If any sentence received over 4, please return yes. Otherwise please return no. Return only yes/no, no explanations or intermediate results.''



\section{Dementia Perception Annotation}
\label{app:perception_annotation}

\begin{table}[t]
\resizebox{\columnwidth}{!}{%
\begin{tabular}{ll}
\toprule
\textbf{Variable}                                                                              & \textbf{Value}                                       \\ 
\midrule
\textbf{N}                                                                                     & 27                                                   \\ \midrule
\multirow{2}{*}{\textbf{Sex}}                                                                  & Female: 16 (59.3\%)                                  \\
& Male: 11 (40.7\%)                                    \\ \midrule
\textbf{Age (years)}                                                                           & 26.6 ± 3.56                                          \\ \midrule
\textbf{Years of education}                                                                    & 16.55 ± 1.82                                         \\ \midrule
\multirow{3}{*}{\textbf{Mother Tongue}}                                                        & Hebrew: 25 (92.6\%)                                  \\ 
& Russian: 1 (3.7\%)                                   \\ 
& Spanish: 1 (3.7\%)                                   \\ \midrule
\multirow{3}{*}{\textbf{\begin{tabular}[c]{@{}l@{}}Familiarity with \\ dementia\end{tabular}}} & (A) None: 5 (18.5\%)                                 \\ 
& (B) Basic awareness via friends/media: 11 (40.7\%)   \\ 
& (C) Personal interaction with relatives: 11 (40.7\%) \\ \bottomrule
\end{tabular}%
}
\caption{Participant Demographics and Background.}
\label{tab:human_annotators}
\end{table}

\subsection{Human Annotators}

\subsubsection{Annotation Guidelines}
\label{app:annotation_guidelines}

``You will be presented with 180 descriptions of the Cookie Theft Picture (figure \ref{fig:cookie} attached).

\begin{figure*}[t]
    \centering
    \includegraphics[width=1\linewidth]{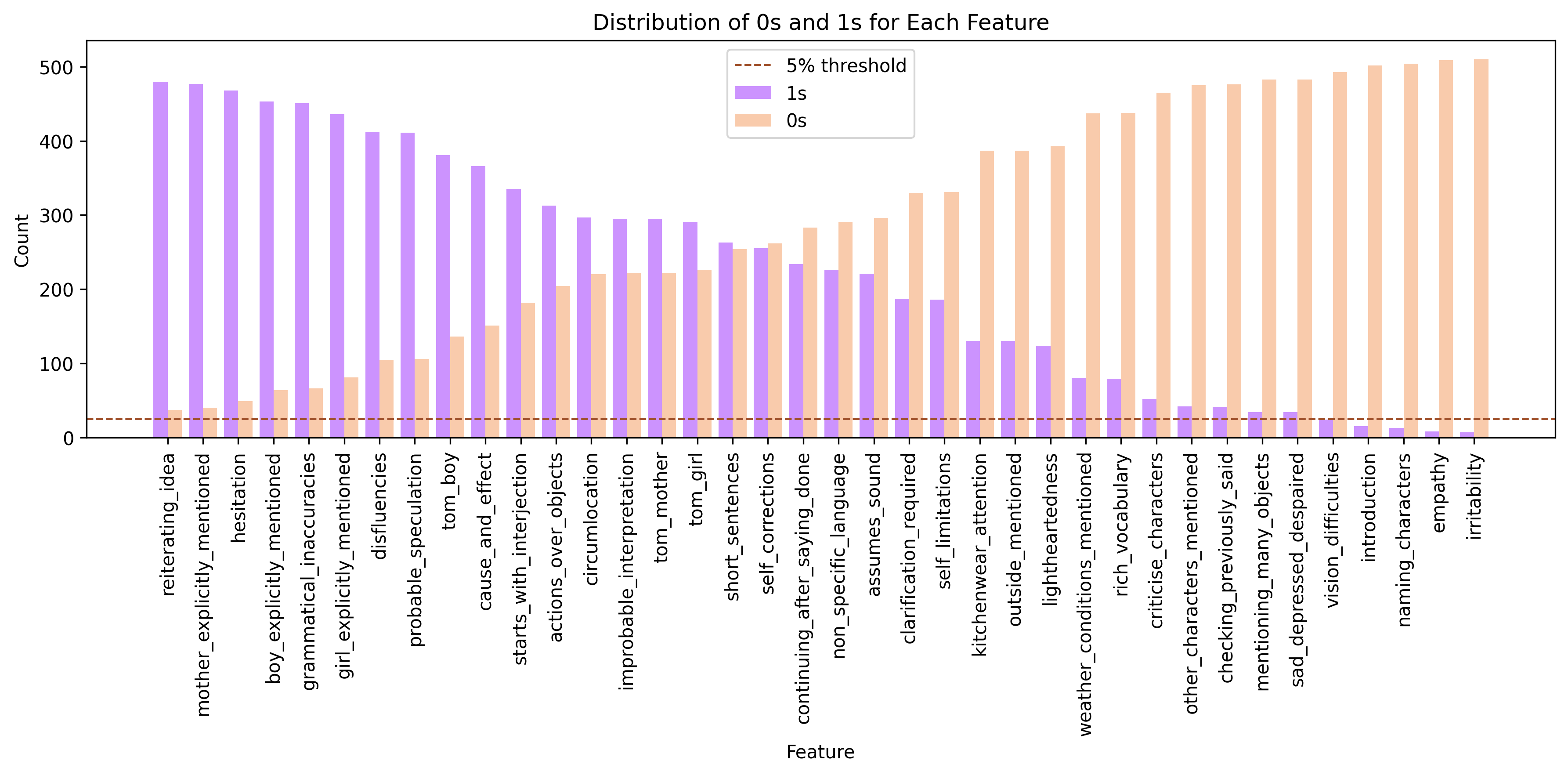}
    \caption{Value distribution across features. Features marked as ‘yes’ in fewer than 5\% of samples (indicated by the dotted line) were excluded from our analyses.}
    \label{fig_app:feature_dist}
\end{figure*}

The speakers (either healthy or with dementia) were asked to describe the picture with as many details as they can. Their responses were recorded, then transcribed from voice to text. The transcripts may include repetitions, incomplete words, non-verbal gestures (laughter, cough…) etc.
For each transcript you read, state your intuition: Does this text feel like it was spoken by a Dementia patient, or a Healthy Control?  
Given what you know / heard / imagine about Dementia / Alzheimer’s patients, please mark ``0'' if this description sounds like it’s from a healthy person, or ``1'' if you sense it came from a dementia patient. 

There is no right or wrong- we are interested in your instincts. We are examining how cognitive decline is perceived by the general population. This can influence people to notice warning signs in their loved ones, prompting them to seek medical attention. It can also reveal biases around the disease. Remember, we are not asking for a clinical diagnosis- just your opinion. Good Luck!''

\subsubsection{Human Annotator Demographics}
\cref{tab:human_annotators} presents the full demographic details of our human annotators. The group includes 16 females and 11 males, aged 22-36. Most are enrolled students (BSc, MSc, or PhD), and one is a postdoctoral researcher. None are native English speakers, though all have knowledge of English as a second or third language and rate themselves as proficient. None of the participants have prior experience as caregivers for individuals with dementia. 

\subsubsection{Recruitment and Consent Procedure}
Permission to recruit annotators was granted by the author's academic institution (IRB). Annotators were recruited through an ad distributed in student WhatsApp groups. A trained member of the research team conducted individual phone conversations to explain the task, during which verbal informed consent was obtained. Following consent, annotators received the data and annotation guidelines. Upon completion of the task, they were compensated with \$50 or academic credit.

\subsection{LLM Perception Prompt}
\label{app:llm_perception_prompt}

``You are analyzing a spoken description of the Cookie Theft Picture from the Boston Diagnostic Aphasia Examination (BDAE). The speakers (either Dementia patients or Healthy Controls) were asked to describe the picture with as many details and/or actions as they can.
Their responses were recorded, then transcribed from voice to text. The transcripts may include repetitions, incomplete words, non-verbal gestures (laughter, cough, etc).
Given what you know / heard / imagine about Dementia / Alzheimer’s patients, please mark whether this description sounds like it’s from a healthy person, or from a Dementia patient. Please use your best judgment.  Remember, we are not asking for a clinical diagnosis- just your perception. Please return ``dementia'' if dementia, ``healthy'' if healthy. Please return ``dementia'' or ``healthy'' only, no explanations or intermediate calculations.''

\section{Data Analysis}
\label{app:data_analysis}

\subsection{Feature Distribution}
\label{app:feature_dist}

\cref{fig_app:feature_dist} presents the number of occurrences of ``1'' or ``0'' for each feature. Five features were marked as positive in fewer than 5\% of the samples (i.e., fewer than 25 samples): \textit{vision difficulties, introduction, naming characters, empathy}, and \textit{irritability}, and were therefore cleaned from our dataset and disregarded throughout our analyses.

\begin{figure*}[t]
    \centering
    \includegraphics[width=1\linewidth]{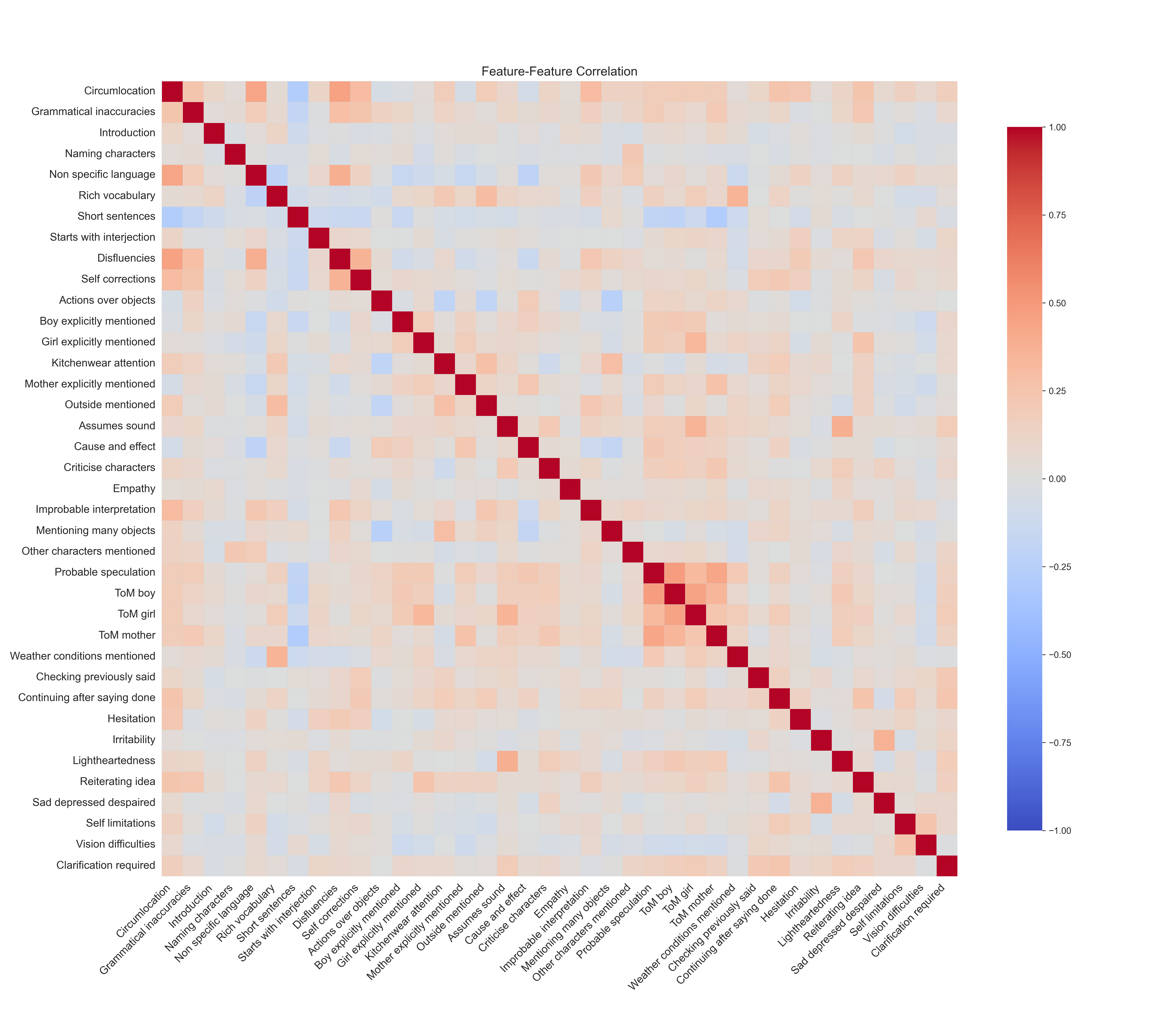}
    \caption{Heatmap of feature correlations (Pearson): red indicates positive correlation, and blue indicates negative.}
    \label{fig:feature_feature_cor}
    \vspace{-0.5em}
\end{figure*}

\begin{figure*}[t]
    \centering
    \includegraphics[width=1\linewidth]{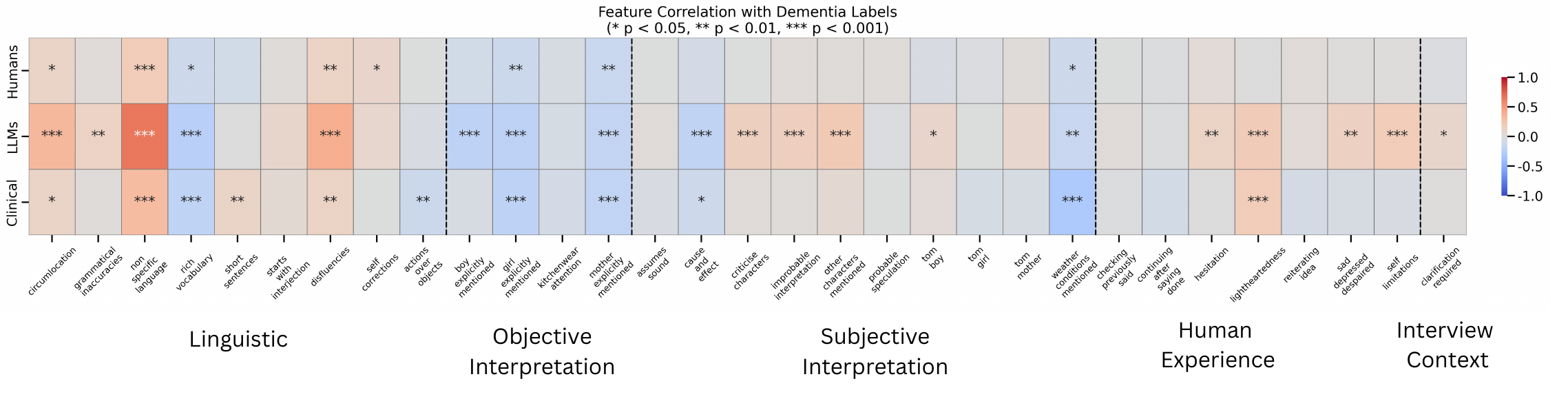}
    \caption{Heatmap of Pearson correlations between features and Human perceptions, LLM majority-vote perceptions, and clinical diagnoses. Red indicates positive correlation with the 'Dementia' label; blue with the 'Healthy' label.}
    \label{fig:heatmap}
    \vspace{-0.5em}
\end{figure*}

\subsection{Feature-Feature Correlations}

\cref{fig:feature_feature_cor} presents Pearson correlations within our 38 features. No extreme correlations are observed, although the Theory of Mind features (ToM boy, ToM girl, and ToM mother) appear somewhat correlated with one another, as well as with probable speculation. This pattern is intuitive: probable speculation, i.e., interpreting something not indisputably visible in the picture, may overlap with describing emotions or motivations. Also, a tendency to apply ToM to one character may reflect an inclination to do so across multiple characters.

\subsection{Feature-Judgment Correlations}

Figure~\ref{fig:heatmap} presents Pearson correlations between our 38 features and the three judgment types: human perception (top row), LLM perception (middle row), and clinical diagnosis (bottom row). Linguistic features (\textit{disfluencies, non-specific language}, etc.) consistently correlate with the “Dementia” label across all three judgments. LLMs show correlations with a broader range of cues than either humans or clinicians, correlating with Human Experience features like \textit{lightheartedness} and \textit{negative sentiment}. Additionally, they show strong associations (p < 0.001) with 13 features and additional correlations (p < 0.05) with 6 more-- nearly double those seen in human perception. LLM judgments also correlate with all five feature categories, suggesting they draw on a wide spectrum of cues. In contrast, human perception is narrower: linguistic cues suggest dementia, while Objective Interpretation features such as mentioning the girl, mother, or outdoors align with perceived healthiness. Clinical diagnosis aligns most with Linguistic and Objective features, with some sensitivity to Subjective elements like weather mentions. Interestingly, weather, although not shown in the image, is significantly correlated across all three sources, making it the only Subjective Interpretation feature shared in this way. Clinical diagnosis also identifies short sentences as a dementia marker, consistent with prior findings \cite{hierLanguageDisintegrationDementia1985, blankenSpontaneousSpeechSenile1987, murrayDistinguishingClinicalDepression2010, liCuriousCaseRetrogenesis2024, agmonAutomatedMeasuresSyntactic2024}. Finally, clinical judgments show fewer correlations with emotional and behavioral cues compared to LLMs, likely because clinicians rely on multimodal signals (tone, expressions, body language) during in-person interviews, cues that LLMs must infer from language alone.

\section{Perception Modeling- Sup. Material}
\label{app:prediction_metrics}

\subsection{All Significant Coefficients}

\cref{tab:coefficients} shows all significant coefficients for the ``Dementia'' and ``Healthy'' labels in the human perception, LLM perception, and clinical diagnosis models. Positive coefficients indicate association with ``Dementia'' while negative coefficients indicate association with ``Healthy''. These results are also visualized in \cref{fig:main_triplet}.

\begin{table*}[t]
\centering
\resizebox{\textwidth}{!}{%
\begin{tabular}{l|l|rrrll}
\toprule
\rowcolor[HTML]{EFEFEF} 
Judgment                                                    & Feature Name                 & $\beta$      & SE    & Wald $Z$
 & $p$-value
            & Significance \\ 
\midrule
\cellcolor[HTML]{EDAD94}                                     & weather conditions mentioned & -1.567 & 0.332 & -4.71  & $p$ \textless{} 0.0001 & ***          \\  
\cellcolor[HTML]{EDAD94}                                     & non specific language        & 1.091  & 0.217 & 5.024  & $p$ \textless{} 0.0001 & ***          \\
\cellcolor[HTML]{EDAD94}                                     & girl explicitly mentioned    & -1.165 & 0.329 & -3.542 & 0.0003             & ***          \\ 
\cellcolor[HTML]{EDAD94}                                     & lightheartedness             & 0.934  & 0.26  & 3.592  & 0.0003             & ***          \\

\cellcolor[HTML]{EDAD94}                                     & mother explicitly mentioned  & -1.467 & 0.528 & -2.778 & 0.0054             & **           \\ 
\cellcolor[HTML]{EDAD94}                                     & short sentences              & 0.598  & 0.215 & 2.783  & 0.0053             & **           \\ 

\cellcolor[HTML]{EDAD94}                                     & actions over objects         & -0.589 & 0.216 & -2.72  & 0.0065             & **           \\ 

\multirow{-8}{*}{\cellcolor[HTML]{EDAD94}Clinical Diagnosis} & probable speculation         & 0.724  & 0.275 & 2.63   & 0.0085             & **           \\ 
\midrule
\cellcolor[HTML]{AAB8DA}                                     & non specific language        & 2.957  & 0.318 & 9.278  & $p$ \textless{} 0.0001 & ***          \\ 
\cellcolor[HTML]{AAB8DA}                                     & girl explicitly mentioned    & -1.821 & 0.439 & -4.141 & $p$ \textless{} 0.0001 & ***          \\ 
\cellcolor[HTML]{AAB8DA}                                     & outside mentioned            & -1.467 & 0.371 & -3.953 & $p$ \textless{} 0.0001 & ***          \\ 
\cellcolor[HTML]{AAB8DA}                                     & disfluencies                 & 2.166  & 0.577 & 3.754  & 0.0001             & ***          \\ 

\cellcolor[HTML]{AAB8DA}                                     & criticise characters         & 1.811  & 0.494 & 3.662  & 0.0002             & ***          \\ 
\cellcolor[HTML]{AAB8DA}                                     & rich vocabulary              & -1.467 & 0.496 & -2.956 & 0.0031             & **           \\ 
\cellcolor[HTML]{AAB8DA}                                     & boy explicitly mentioned     & -1.537 & 0.466 & -3.299 & 0.0009             & ***          \\ \cellcolor[HTML]{AAB8DA}                                     & lightheartedness             & 0.829  & 0.338 & 2.446  & 0.0144             & *            \\ 
\cellcolor[HTML]{AAB8DA}                                     & self limitations             & 0.697  & 0.289 & 2.409  & 0.0159             & *            \\
\cellcolor[HTML]{AAB8DA}                                     & actions over objects         & -0.76  & 0.317 & -2.393 & 0.0166             & *            \\ 
\cellcolor[HTML]{AAB8DA}                                     & sad depressed despaired      & 1.252  & 0.545 & 2.298  & 0.0215             & *            \\ 
\cellcolor[HTML]{AAB8DA}                                     & tom boy                      & 0.839  & 0.374 & 2.241  & 0.025              & *            \\ 
\multirow{-13}{*}{\cellcolor[HTML]{AAB8DA}LLM Perception}    & other characters mentioned   & 1.065  & 0.527 & 2.022  & 0.0431             & *            \\ 
\midrule
\cellcolor[HTML]{93C7B7}                                     & non specific language        & 0.568  & 0.19  & 2.99   & 0.0027             & **           \\ 
\cellcolor[HTML]{93C7B7}                                     & girl explicitly mentioned    & -0.572 & 0.263 & -2.17  & 0.0299             & *            \\ 
\cellcolor[HTML]{93C7B7}                                     & short sentences              & -0.571 & 0.192 & -2.965 & 0.003              & **           \\ 
\multirow{-4}{*}{\cellcolor[HTML]{93C7B7}Human Perception}   & mother explicitly mentioned  & -0.758 & 0.372 & -2.037 & 0.0416             & *            \\ \bottomrule
\end{tabular}%
}
\caption{Coefficient analysis for logistic regression models predicting each judgment type. The regression coefficient is denoted by $\beta$, with larger absolute values indicating a stronger influence. Positive $\beta$ values suggest an association with the \textit{Dementia} label, and negative values with \textit{Healthy}. \textit{SE} denotes the standard error. The Wald statistic tests whether $\beta = 0$. $p$-values reflect each feature's significance in predicting perceived or diagnosed dementia.}
\label{tab:coefficients}
\end{table*}

\subsection{Prediction Evaluation}
\label{app:prediction_evaluation}

\cref{sec:logreg_modeling} described the stepwise logistic regression models used to analyze coefficients linked to human perception, LLM perception, and clinical diagnoses. Since our primary goal is to interpret the learned coefficients and derive meaningful insights, we initially trained the models using the entire dataset. However, to comply with NLP standards and evaluate predictive performance, we then train models using only the statistically significant features for each judgment group (\cref{fig:main_graph}). For example, to predict human perception, we train a stepwise logistic regression model using only the features found significant for that group. 

We use a 5-fold cross-validation with test folds of 103-104 instances each, reporting average performance on the held-out test sets. To prevent data leakage, we ensure that no patient with multiple samples (due to the longitudinal structure of the Pitt corpus) appears in both training and test sets.

\cref{tab:precision_recall_f1} presents accuracy, precision, recall, F1 score, and ROC-AUC for each model. Performance is notably strong when predicting LLM perception, possibly due to the broader range of features they rely on (13 features, compared to 8 and 4 for clinical diagnosis and human perception, respectively).

Lower scores were observed for the prediction of clinical diagnosis. This is expected, given our approach of representing Cookie Theft picture descriptions using a small number of binary, high-level features. State-of-the-art studies aiming to optimize prediction on the Pitt corpus report accuracies around 85\%; however, they rely on large transformer architectures that are hyperparameter-tuned on the full raw text, and sometimes audio as well \cite{ilias2022explainable}. Studies more similar in nature to ours -- i.e., those that represent Cookie Theft transcripts using relatively few, well-defined extracted features -- report predictive metrics more in line with our results (e.g., F1 scores around 70\%) \cite{sirts2017idea, wankerl2017n}. Accordingly, our regression model for predicting clinical diagnoses, based solely on eight high-level features, is not optimized for high-accuracy dementia detection. However, it is designed to provide insight into how different groups perceive dementia.

Finally, when looking at human perception, it is no surprise that it is difficult to predict. Throughout our work, we have repeatedly demonstrated the inherent inconsistency, subjectivity, and confusion in human judgments; even the annotators themselves lacked clear insight into their decision process.


\begin{table}[t]
\resizebox{\columnwidth}{!}{%
\begin{tabular}{l|c|c|c}
                   \toprule
                   & \cellcolor[HTML]{EDAD94}Clinical & \cellcolor[HTML]{AAB8DA}LLM Perc. & \cellcolor[HTML]{93C7B7}Human Perc. \\
\midrule
\textbf{Accuracy}  & 67.3                                      & 84.1                                  & 60.1                                    \\ 
\textbf{Precision} & 69.9                                      & 80.7                                  & 58.1                                    \\ 
\textbf{Recall}    & 75.4                                      & 81.0                                  & 42.9                                    \\ 
\textbf{F1 Score}  & 71.1                                      & 80.5                                  & 48.7                                    \\ 
\textbf{ROC-AUC}   & 78.4                                      & 92.1                                  & 62.5                                    \\ 
\bottomrule
\end{tabular}%
}
\caption{Evaluation metrics for the predictive performance of logistic regression perception models across all three judgment types.}
\label{tab:precision_recall_f1}
\end{table}

\section{Analysis of Human-provided Rationale}
\label{app:human_rationale}

We received reports from 18 of the 27 annotators, describing the patterns they noticed and believed they relied on to assess whether the speaker appeared cognitively impaired or healthy. We translated these responses into English and manually reviewed them. Raw annotator responses are presented in \cref{app:raw_annotator_responses}. We distilled these responses into concepts, some align with features from our predefined expert-guided list, and others representing new, emergent patterns. We then tallied the number of times each of these concepts/features was mentioned. The results of this analysis are shown in \cref{tab:human_rationale_counts}. The most frequently mentioned features were \textit{reiterating ideas} (10 mentions) and \textit{disfluencies} (7 mentions).

Notably, some concepts introduced by annotators were broader in scope than our predefined categories. Among the emergent concepts, \textit{level of detail} was the most frequently cited  possibly due to its encompassing nature, covering a range of specific content features. Similarly, some annotators referred generally to the attribution of emotions to characters, without identifying which characters. A particularly nuanced observation regarding \textit{level of detail} is that some associated it with signs of cognitive impairment, while other interpreted it as indicative of cognitive intactness. 

Some features also spanned multiple interpretive categories. For example, \textit{confusion} could be understood as a linguistic, interpretive, or contextual cue, and \textit{laughter occurrence} may relate both to the speaker's emotional state and the dynamics of the interview setting.
This suggests that annotators often focused on broader impressions rather than specific, fine-grained elements.

\begin{table}[t]
\centering
\resizebox{\columnwidth}{!}{%
\begin{tabular}{|l|l|l|}
\hline
                                                                                                   & Feature Name                & Num Mentions \\ \hline
\multirow{13}{*}{\begin{tabular}[c]{@{}l@{}}Fully \\ correspond\\ to our \\ features\end{tabular}} & reiterating ideas           & 10           \\ \cline{2-3} 
                                                                                                   & disfluencies                & 7            \\ \cline{2-3} 
                                                                                                   & improbable interpretation   & 5            \\ \cline{2-3} 
                                                                                                   & checking previously said    & 3            \\ \cline{2-3} 
                                                                                                   & circumlocution              & 2            \\ \cline{2-3} 
                                                                                                   & clarification required      & 2            \\ \cline{2-3} 
                                                                                                   & naming characters           & 2            \\ \cline{2-3} 
                                                                                                   & rich vocabulary             & 2            \\ \cline{2-3} 
                                                                                                   & short sentences             & 2            \\ \cline{2-3} 
                                                                                                   & self limitations            & 1            \\ \cline{2-3} 
                                                                                                   & boy explicitly mentioned    & 1            \\ \cline{2-3} 
                                                                                                   & girl explicitly mentioned   & 1            \\ \cline{2-3} 
                                                                                                   & mother explicitly mentioned & 1            \\ \hline
\multirow{7}{*}{\begin{tabular}[c]{@{}l@{}}Additional\\ features\\ mentioned\end{tabular}}         & level of detail             & 9            \\ \cline{2-3} 
                                                                                                   & incoherence                 & 5            \\ \cline{2-3} 
                                                                                                   & total length                & 3            \\ \cline{2-3} 
                                                                                                   & talk about self             & 1            \\ \cline{2-3} 
                                                                                                   & assigning emotions          & 1            \\ \cline{2-3} 
                                                                                                   & laughter                    & 1            \\ \cline{2-3} 
                                                                                                   & confusion                   & 1            \\ \hline
\end{tabular}
}
\caption{Concepts reported by human annotators.}
\label{tab:human_rationale_counts}
\end{table}

\subsection{Raw Annotators Responses}
\label{app:raw_annotator_responses}

\begin{itemize}
    \item Fragmented sentences, repetitions, confusion, and self-questioning like “Did I already say that?”
    \item Marked as “Dementia” if there was reference only to part of the picture, mentions of things not shown, overly short or discontinuous descriptions, many questions, incoherence, or excessive focus on tiny details.
    \item Repetition of the same sentence or phrase, messy or incoherent flow, or exact repetition of a sentence triggered a dementia label.
    \item Explicit self-checks (e.g., “Did I already say that?”) and topic jumps were viewed as dementia indicators.
    \item Sentences that faded mid-way or had abrupt topic changes; exaggerated imagination or personal tangents were seen as red flags.
    \item Initially thought humming might be a sign, but later decided it wasn’t a reliable cue.
    \item Looked for attention patterns, whether patients noticed different details, but found no clear pattern.
    \item Hesitated over emotional expressions like “oh no” or laughter but didn’t use them as indicators.
    \item Often relied on gut feeling: either excessive rambling or overly clipped responses missing key details stood out.
    \item Confusion between characters, fading thoughts (not normal stuttering), assigning names, and projecting emotions/inferences were red flags.
    \item Excessive ellipses or repetitive statements like “I don’t know, I don’t know” were used as indicators.
    \item Less about content and more about delivery, the words themselves.
    \item Looked for repetition, vocabulary level, length of response, and level of detail (with detailed = healthy).
    \item Scattered responses or inclusion of elements not present in the image were marked as dementia.
    \item Naming characters or imagining beyond the image was a red flag.
    \item Repetition and lack of detail were key; detail often equated with cognitive health.
    \item Confused or incorrect responses (e.g., interpreting cookie quantity emotionally) raised suspicion of decline.
    \item Clear general descriptions were often seen as indicative of intact memory.
    \item Noticed differences in attention to detail across participants, some highly detailed, others gave almost nothing.
    \item Repetition spaced out in the text and incoherent sentence flow with emotional inserts were used as dementia cues.
    \item Requests to repeat the task, especially mid-description, were noted as potential signs.
    \item Strange repetition or failure to find words (e.g., calling a stool a chair) stood out.
    \item Incoherent phrasing, incomplete ideas, and inability to retrieve specific words were viewed as markers.
    \item People who laughed were perceived as healthy.
    \item Repetition of actions, lack of detail, or unrelated storytelling triggered a dementia classification.
    \item Jumping between characters inconsistently (e.g., finishing with the children, moving to the mother, then back) was noted.
\end{itemize}

\end{document}